\documentclass[runningheads]{llncs}

\usepackage{eccv}

\usepackage{eccvabbrv}

\usepackage{graphicx}
\usepackage{booktabs}

\usepackage[accsupp]{axessibility}  

\usepackage{hyperref}

\usepackage{orcidlink}

\usepackage{algorithm}
\usepackage{algorithmicx}
\usepackage{algpseudocode}
\usepackage{multirow}
\usepackage{adjustbox}
\usepackage{threeparttable}
\usepackage{pifont}
\usepackage{array}
\usepackage{enumitem}
\usepackage{bbding}
\usepackage{fontawesome}

\usepackage{listings}
\usepackage{graphicx}
\usepackage{booktabs}
\usepackage{colortbl}
\usepackage{url}
\usepackage{mathtools}
\usepackage{microtype}
\usepackage{float}
\usepackage{bm}
\usepackage{adjustbox}
\usepackage{caption}
\usepackage{subcaption}
\usepackage{siunitx}  
\usepackage{makecell} 
\usepackage{wrapfig}

\definecolor{lightgray}{rgb}{0.9, 0.9, 0.9}
\definecolor{lightblue}{rgb}{0.8, 0.9, 1.0}   
\definecolor{beige}{rgb}{0.96, 0.96, 0.86}    
\definecolor{lightgreen}{rgb}{0.9, 1.0, 0.9}  
\definecolor{lavender}{rgb}{0.9, 0.9, 0.98}   
\definecolor{lightyellow}{rgb}{1.0, 1.0, 0.88} 
\definecolor{brightgreen}{RGB}{102, 255, 0} 
\definecolor{deepgreen}{RGB}{50, 186, 70} 

\newcommand{\rowstyle}[1]{\gdef\currentrowstyle{#1}%
  \aftergroup\currentrowstyle}

\begin{document}

\title{Adversarial Prompt Tuning for Vision-Language Models} 


\author{Jiaming Zhang\inst{1}\thanks{This work is done when the author interned at Fudan University.} \and
Xingjun Ma\inst{2} \and
Xin Wang\inst{2} \and
Lingyu Qiu\inst{3} \and
Jiaqi Wang\inst{1} \and
Yu-Gang Jiang\inst{2} \and
Jitao Sang\inst{1}
}

\authorrunning{J.~Zhang et al.}

\institute{Beijing Jiaotong Univisity, Beijing, China \\
\and
Fudan Univisity, Shanghai, China\\
\and
Nanjing University of Aeronautics and Astronautics, Nanjing, China\\
}

\maketitle

\begin{abstract}
  With the rapid advancement of multimodal learning, pre-trained Vision-Language Models (VLMs) such as CLIP have demonstrated remarkable capacities in bridging the gap between visual and language modalities. 
However, these models remain vulnerable to adversarial attacks, particularly in the image modality, presenting considerable security risks. 
This paper introduces \textbf{Adversarial Prompt Tuning (AdvPT)}, a novel technique to enhance the adversarial robustness of image encoders in VLMs. 
AdvPT innovatively leverages learnable text prompts and aligns them with adversarial image embeddings, to address the vulnerabilities inherent in VLMs without the need for extensive parameter training or modification of the model architecture. 
We demonstrate that AdvPT improves resistance against white-box and black-box adversarial attacks and exhibits a synergistic effect when combined with existing input denoising defense techniques, further boosting defensive capabilities. 
Comprehensive experimental analyses provide insights into adversarial prompt tuning, a novel paradigm devoted to improving resistance to adversarial images through textual input modifications, paving the way for future robust multimodal learning research. 
These findings open up new possibilities for enhancing the security of VLMs. 
Our code is available at \url{https://github.com/jiamingzhang94/Adversarial-Prompt-Tuning}.
\let\thefootnote\relax\footnotetext{Corresponding authors: Xingjun Ma and Jitao Sang.}
\end{abstract}

\section{Introduction}
\label{sec:intro}

Large-scale pre-trained Vision-Language Models (VLMs) have demonstrated superb capabilities in generalizing to a wide variety of downstream tasks. 
These architectures are trained to bridge the gap between visual and language modalities, as demonstrated by the huge amount of web-scale data~\cite{radford2021learning}. 
With the increasing trend of multimodal learning, there is a growing number of VLMs being released to the public, leading to rapid growth of downstream applications.
However, many studies have revealed that VLMs, similar to traditional visual models, are also vulnerable to small adversarial noises, which is a major security threat to deep neural networks (DNNs)~\cite{szegedy2013intriguing, zhou2023advclip}. 
In particular, noise in the image modality is markedly more invisible compared to token replacement in the text modality. 
Therefore, the imperative task of enhancing the adversarial robustness of the image encoders in VLMs requires an effective solution.

\begin{wrapfigure}{r}{0.5\textwidth} 
  \centering
  \includegraphics[width=0.48\textwidth]{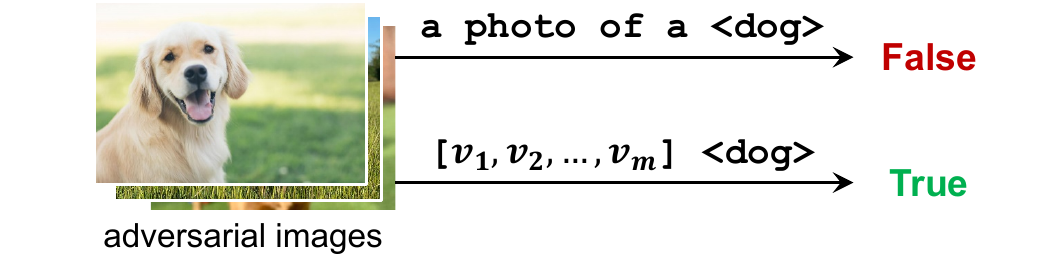} 
  \caption{The defending effect of \emph{AdvPT}: hand-crafted prompts (top) fail to match with adversarial images, whereas prompts constituted by learnable vectors (bottom) enable correct recognition.}
  \label{fig:1}
\end{wrapfigure}

In the image domain, adversarial training (AT) has been proven to be the most effective approach for training robust DNNs against adversarial examples (inputs with adversarial noise)~\cite{madry2017towards}. 
AT is usually formulated as a min-max optimization problem, which generates adversarial examples at each training iteration to update the image encoder. Therefore, it is computationally expensive, and thus cannot be easily applied to train large VLMs.
As such, recent works turn to input pre-processing techniques like diffusion-based purification to improve the adversarial robustness of VLMs~\cite{xie2018mitigating, mustafa2019image, nie2022diffusion}.

Drawing from advances in Natural Language Processing (NLP), we observe a shift from fixed text prompts, such as ``\texttt{a photo of a <category>}'',  to learnable prompts in CLIP's text encoder~\cite{zhou2022learning, zhou2022conditional}. 
Such a transition can help enhance image-text matching. 
Inspired by the idea of prompt tuning, in this work we propose \textbf{\emph{Adversarial Prompt Tuning}} (\textbf{\emph{AdvPT}}), a novel approach that improves the adversarial robustness of the image encoders in VLMs using learnable prompts.
As depicted in \cref{fig:1} and \cref{fig:framework}, \emph{AdvPT} models the textual prompt with learnable vectors and aligns the clean text embedding with adversarial image embedding to improve adversarial robustness.
Specifically, we generate the adversarial images using the image encoder and then compute and save the embeddings of the adversarial images into an adversarial embedding bank.  We then discard the image encoder but use the adversarial embedding bank to enhance the adversarial robustness, i.e., we align the clean text embedding with the adversarial image embedding through prompt tuning.
This process involves gradient backpropagation through the text encoder to optimize the learnable vectors while preserving the pre-trained parameters. In a nutshell, our \emph{AdvPT} leverages the text encoder’s inherent knowledge for rectifying the adversarial embeddings (pre-computed with the image encoder).

We evaluate \emph{AdvPT} against both white-box and black-box adversarial attacks on 8 image datasets, and show that it outperforms the vanilla CLIP (with hand-crafted prompts) by a considerable margin. 
By focusing on textual input processing and alignment, \emph{AdvPT} opens up a new direction for augmenting adversarial robustness in VLMs. 
It can also be integrated with image-based defense strategies to further boost the adversarial robustness of the image modality.
We also observe a generalization-robustness trade-off in \emph{AdvPT}, similar to that in traditional AT.
We further evaluate the domain transferability of the learnable vectors, testing their performance across various datasets after training on a specific one.
Lastly, we conduct an in-depth analysis of the learned vectors and reveal the closest words associated with the vectors, gaining more understanding of the working mechanism of \emph{AdvPT}.

In summary, our main contributions are:
\begin{itemize}
    \item  We propose a novel method \emph{Adversarial Prompt Tuning} (\emph{AdvPT}) to enhance the adversarial robustness of VLMs by aligning the text embeddings with adversarial image embeddings. Specifically, we robustify the image encoder in VLM against adversarial examples using textual prompt modifications.
    \item We demonstrate the effectiveness of \emph{AdvPT} on various image datasets, showing its superiority over the vanilla CLIP. It can also be combined with input purification methods to further boost the robustness. 

    \item We also provide a set of understandings of the working mechanism of \emph{AdvPT}, the generalization-robustness trade-off, the adaptability of the learned vectors to domain shift, and their linguistic meanings. These understandings can help guide future work to leverage textual input to counter adversarial images.
\end{itemize}

\section{Related Work}
\label{sec:releted}

\subsection{Vision Language Models}

VLMs have achieved remarkable success and demonstrated superb capabilities across a wide range of tasks.
These models are typically classified into two groups.
The first is grounded in large NLP models enhanced with visual modality capabilities, exemplified by GPT-4V~\cite{gpt}.
The second group, represented by CLIP~\cite{radford2021learning} and ALIGN~\cite{jia2021scaling}, treats image and language modalities with equal emphasis.
These models acquire joint image-language representations through self-supervised learning from vast data pools.
Our study focuses on the latter category of VLMs, specifically on improving the adversarial robustness of their image encoders for image recognition tasks.

\subsection{Prompt Learning}

The concept of prompt learning originated in the field of NLP. 
It refers to fine-tuning the prompts instead of model parameters (freezing the model).
Research in prompt learning aims to automatically learn more effective prompts instead of using a hand-crafted prompt~\cite{li2021prefix, liu2021p}.
This approach has been extended to visual models~\cite{jia2022visual, wang2022learning} and vision-language models~\cite{zhou2022learning, zhou2022conditional, khattak2023maple}, with the unified objective of enhancing model accuracy through prompt refinement.
Our study, while grounded in the CoOp framework~\cite{zhou2022learning}, diverges in its objective. CoOp represents the initial foray into prompt learning within the visual-language domain, distinguished by its simplicity and rapid processing pipeline.
Instead of improving image recognition performance, our focus shifts to leveraging textual input modifications to improve the adversarial robustness of the image encoder.

\subsection{Adversarial Defenses}

Combating against adversarial images remains an unresolved challenge. 
Adversarial defenses broadly fall into two camps: model robustification methods and input denoising methods.
The former includes methods like AT~\cite{madry2017towards}, Fast AT~\cite{wongfast2020}, TRADES~\cite{zhang2019theoretically} and MART~\cite{wang2019improving}.
This methodology is usually expressed as a min-max optimization problem, with continuous updates to the model parameters across all training iterations. 
However, this process is computationally demanding, posing difficulties for deployment on VLMs due to the scale of the model and dataset.
As a result, the latter approach based on the image process has emerged as a solution suited for VLMs.

The adversarial defense through input image modification is straightforward in its essence. 
It removes or weakens the impact of adversarial noise through inference-time methods such as input transformations~\cite{guo2017countering, mustafa2019image}, smoothing~\cite{liao2018defense, salman2020denoised, zhangrobustify}, and rescaling~\cite{xie2018mitigating}.
For example, Xie \emph{et al.}~\cite{xie2018mitigating} employed random image rescaling to diminish adversarial effects, and Mustafa \emph{et al.}~\cite{mustafa2019image} utilized image super-resolution as a defense mechanism.
Although somewhat limited in efficacy, these methods are pragmatically valuable for their efficiency. 
Recently, adversarial purification based on diffusion models has emerged~\cite{yoon2021adversarial, nie2022diffusion}. 
Nie \emph{et al.}\cite{nie2022diffusion} introduced the powerful adversarial purification, DiffPure, to address the shortcomings of previous approaches, albeit with increased time complexity.
Mao \emph{et al.}\cite{mao2022understanding} identifies that AT of the CLIP on one dataset struggles to impact another dataset, defining this as the zero-shot adversarial robustness problem, and introduced visual prompt tuning~\cite{jia2022visual} to address this.

Our approach deviates from these strategies by not modifying the model nor the input image, presenting a novel defense mechanism against adversarial images. The subsequent sections detail our method and its integration with existing defensive techniques.

\section{Revisiting Clip and the Adversarial Robustness of Its Image Encoder}
\label{sec:revisiting}
\subsection{CLIP}
We provide a concise introduction to VLMs, with an emphasis on the CLIP architecture. 
While our methods are tailored to CLIP, they are potentially extendable to a broader range of VLMs within the contrastive learning framework.

CLIP comprises two distinct encoders: one for images and the other for text. The image encoder aims to distill image embeddings from the input visuals, utilizing either a Convolutional Neural Network (CNN)~\cite{he2016deep} or a Vision Transformer (ViT)~\cite{dosovitskiy2020image} backbone.
In contrast, the text encoder relies on a Transformer~\cite{vaswani2017attention} to generate embeddings from textual data.

During its training phase, CLIP leverages contrastive loss to develop a unified embedding space between visual and language modalities.
Upon completion of training, CLIP finds utility in zero-shot image recognition, facilitated through an image-text retrieval mechanism.
For example, in the prompt ``\texttt{a photo of a <class>}'', replacing \texttt{<class>} with specific categories from a dataset with $K$ classes allows the model to assess the similarity between an image and $K$ textual descriptions.

Denoting input images as $x$ and their corresponding image embeddings from encoder $E(\cdot)$ as $e$, and considering a set of textual prompts $\{w_i\}_{i=1}^{K}$ as text embeddings produced by text encoder $G(\cdot)$, the prediction probability is mathematically expressed as follows:
\begin{equation}\label{eq:1}
    p(y|e) = \frac{\exp(\text{sim}(e, w_y) / \tau)}{\sum_{i=1}^{K} \exp(\text{sim}(e, w_i) / \tau)},
\end{equation}
where $\rm{sim}(\cdot, \cdot)$ denotes cosine similarity with a temperature parameter $\tau$.

\subsection{Adversarial Robustness of CLIP’s Image Encoder}

We first introduce our threat model, which describes the assumed knowledge of the adversary, from what inputs they can manipulate to their access to the model architecture and parameters.
Our study focuses on the adversarial robustness of image encoders, assuming that the attacker has full knowledge of the model architecture and parameters of image and text encoders, and can perturb the image input.
\emph{However, the adversary has no control over the textual input nor knowledge of prompt tuning.}
Therefore, text adversarial attacks are also not applicable here~\cite{zhao2023evaluating, zhou2023advclip, zhang2022towards}.

We now introduce the adversarial attacks that target the image encoders.
Consider an original input image \( x \), with \( \delta \) symbolizing adversarial noise. The adversarial example \( x' = x + \delta \), once processed by the image encoder \( E \), generates an adversarial embedding \( e' \).

Adversaries can employ two objective functions to impair the accuracy of matching with textual descriptions. 
The first objective is to make the adversarial embedding \( e' \) markedly diverge from the embedding \( e \) of the original image, i.e., to maximize the discrepancy between \( e' \) and \( e \). 
The second objective is to ensure the adversarial embedding \( e' \) does not align with the corresponding ground-truth textual description embedding \( w_g\), i.e., to maximize the discrepancy between \( e' \) and \( w_g\).
PGD and AutoAttack are deployed to represent the former and latter objectives, respectively.
In this work, we focus on \( \ell_\infty \)-norm constrained perturbations, where each \( \delta \) adheres to \( \| \delta \|_\infty \leq \epsilon \), with \( \epsilon \) denoting the maximum allowable perturbation magnitude.

To defend against adversarial images, existing defense methods generally fall into two categories: model robustification methods and input denoising methods.
As mentioned above, model robustification methods like AT struggle to handle VLMs due to efficiency issues.
The input denoising operation can be conceptualized as a function \( h \) that processes adversarial images, aiming to minimize the disparity between \( E(h(x')) \) and \( E(x) \).

\section{Adversarial Prompt Tuning}

\begin{figure*}[t]
\centering
\includegraphics[width=\textwidth]{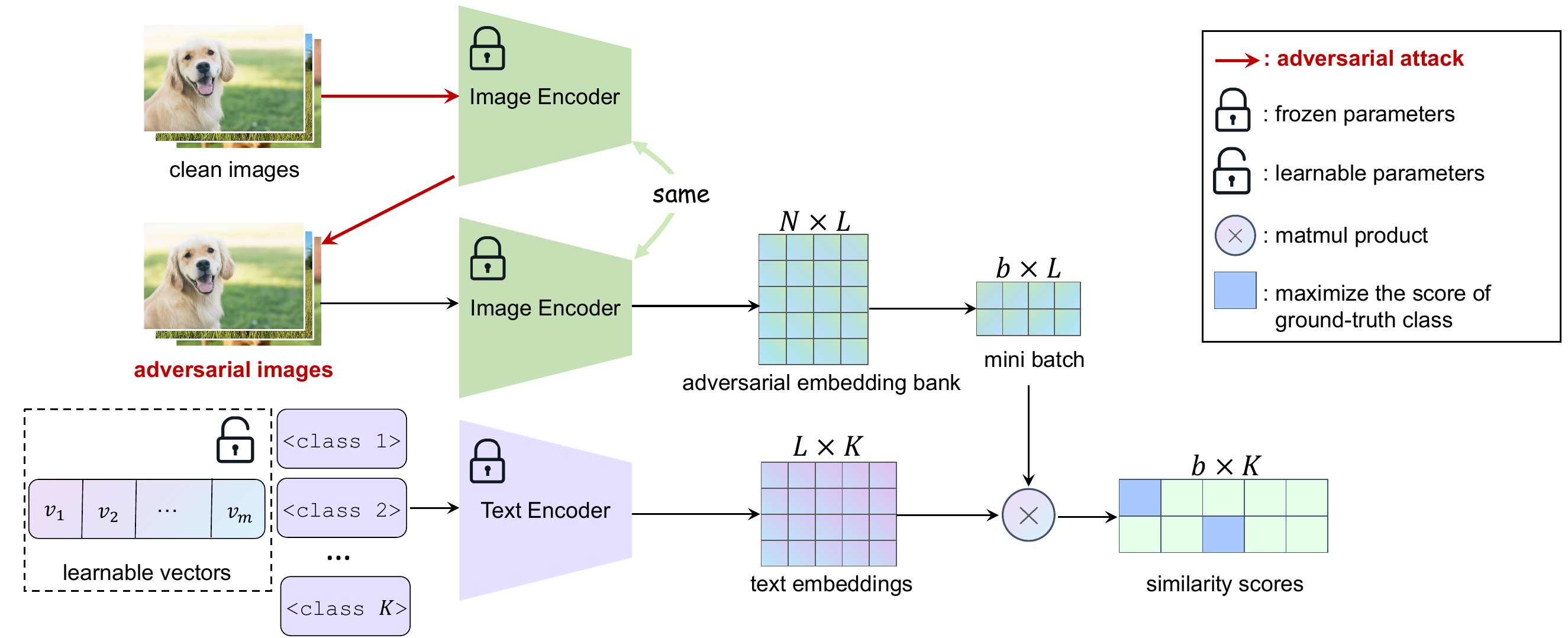} 
\caption{An overview of the \emph{AdvPT} framework.} 
\label{fig:framework} 
\end{figure*}

\paragraph{Overview.}
Our proposed method, \emph{AdvPT}, involves optimizing learnable vectors as text prompts to enhance the robustness against image adversarial attacks.
This diverges from previous context optimization approaches~\cite{zhou2022learning, zhou2022conditional} aimed at increasing image recognition rates.
\cref{fig:framework} provides an framework overview of \emph{AdvPT}.
On a $K$-class dataset \(D=\{(x_i, y_i)\}_{i=1}^{N}\) of $N$ images and corresponding texts, \emph{AdvPT} begins with feeding the clean images \(x\) into the image encoder $E$ to generate its adversarial image \(x'\).
The adversarial images are then fed into the image encoder $E$ to obtain the adversarial image embeddings into an adversarial embedding bank $\mathbf{A} \in \mathbb{R}^{N \times L}$, where $L$ is the embedding dimension. 
The image encoder $E$ is discarded in the subsequent steps.
This approach is entirely distinct from traditional defensive methods, which, whether through augmentation (e.g., visual prompt tuning~\cite{jia2022visual, mao2022understanding}) or modification (e.g., AT~\cite{madry2017towards}) of the parameters of CLIP's image encoder branch, rely on \textbf{on-the-fly adversarial example generation} during each training epoch.
Even with partial parameter tuning (visual prompt), the adversarial example generation necessitates \textbf{complete forward and backward propagation} of gradients through the image encoder, resulting in an untenable burden in the context of VLMs.

On the textual side, the prompt for class $i$ is denoted as \([v_1, v_2, \dots, v_M, c_i]\), with \(c_i\) is the embedding representation of the class name.
These prompts are then processed by the text encoder $G$ to generate text embeddings $\mathbf{T} \in \mathbb{R}^{L \times K}$. During the fine-tuning process, a mini-batch $\mathbf{B} \in \mathbb{R}^{b \times L}$ with batch size $b$ from $\mathbf{A}$ is used to compute the similarity score $\mathbf{S}=\mathbf{B}\mathbf{T} \in \mathbb{R}^{b \times K}$. The objective is to maximize the score of the ground-truth class by optimizing the learnable vectors $V = [v_1, v_2, \dots, v_M]$, through backpropagation in the text encoder.
Overall, the entire process can be roughly divided into two steps: adversarial embedding bank generation and learnable vector optimization. 
Next, we will introduce the two steps in detail.

\subsection{Adversarial Embedding Bank Generation}

To improve the image encoder $E$'s adversarial robustness, \emph{AdvPT} first generates adversarial images on encoder $E$, then re-feeds them into the encoder to obtain and store their adversarial embeddings. Note that \emph{AdvPT} differs greatly from AT, which iteratively generates adversarial examples at each iteration of training and continuously updates the target model on the generated adversarial examples, leading to significant computational costs. 
Conversely, \emph{AdvPT} fixes the parameters of the image encoder $E$, channeling focus exclusively on updating the learnable vectors at the input of the text encoder \(G\).
This strategy significantly diminishes the number of learnable parameters.
With the image encoder $E$ frozen, the generation of the adversarial examples is only a one-pass process.
These examples, once processed through \(E\), constitute the adversarial embedding bank $\mathbf{A}$.
After this step, the image encoder \(E\) is discarded, leaving only the adversarial embedding bank $\mathbf{A}$ for the subsequent prompt tuning.

We employ the PGD attack~\cite{madry2017towards} to generate adversarial images $x'$ on the image encoder \( E(\cdot) \) with $\theta$. This process can be formulated as:
\begin{equation}\label{eq:2}
    x' = x'_{(t+1)} = \Pi_{x+\Omega}(x'_{(t)} + \alpha \cdot \text{sign}(\nabla_{x} J(\theta; x'_{(t)}, x)),
\end{equation}
where $x'{(t)}$ represents the adversarial example at iteration $t$. 
$\Pi$ is the projection. $\Omega$ is the feasible region of $\delta$, which ensures that the perturbed example remains within the allowed limits $\epsilon$. 
$\alpha$ is the step size for each iteration. 
$\nabla_{x} J(\theta; x'_{(t)}, x)$ computes the gradient of the loss function \(J\) with respect to the parameters $\theta$ of $E$, wherein \(J\) serves as a distance metric quantifying the discrepancy in embeddings between \(e' = E(x')\) and \(e = E(x)\).
In our research, we utilize the Kullback-Leibler Divergence, as in TRADES~\cite{zhang2019theoretically}, to serve as our adversarial loss function.

The design of the adversarial embedding bank presents significant advantages. Primarily, it eliminates the need for redundant forward and backward passes through the image encoder, thereby greatly saving computational time. Moreover, the embedding space's lower dimensionality compared to the original image space substantially reduces the required computational memory.

\subsection{Learnable Vector Optimization}

\begin{algorithm}[t]
\caption{Adversarial Prompt Tuning Pipeline}
\label{algorithm}
\begin{algorithmic}[1]
\State {\bfseries Input:} image encoder $E$, text encoder $G$, images $x$ and class name $c$, perturbation restriction $\epsilon$, iteration $t$
\State {\bfseries Output:} learnable vectors $[v_1, v_2, \dots, v_M]$

\State $x' = \text{attack}(x, \epsilon, t; E)$
\State $\mathbf{A} = E(x')$
\State Initialize learnable vectors $V = [v_1, v_2, \dots, v_M]$

\For{$\mathbf{B}$ {\bfseries in} iter$(\mathbf{A})$}
    \State Initialize $\theta_i$
    \State $\mathbf{T} = G([V, c])$
    \State $\mathbf{S}$ = $\mathbf{B}\mathbf{T}$
    \State Optimize $V \leftarrow$ Maximize $\mathbf{S}$
\EndFor
\end{algorithmic}
\end{algorithm}

The next phase in \emph{AdvPT} involves the construction and optimization of the learnable vectors.
Specifically, our method seeks to model textual prompts with learnable vectors \( V = [v_1, v_2, \ldots, v_M] \), optimized by aligning them with adversarial embeddings, thus rectifying the non-robust features of the images utilized by the model.
Initially, the text prompts \( [v_1, v_2, \ldots, v_M, c_i]_{i=1}^{K} \) are fed into the text encoder $G$, producing text embeddings $\mathbf{T} = [w_{1}, w_{2}, \dots, w_{K}]^T \in \mathbb{R}^{L \times K}$.
In the fine-tuning phase, each iteration retrieves a mini-batch $\mathbf{B} = [e'_{1}, e'_{2}, \dots, e'_{b}] \in \mathbb{R}^{b \times L}$ from the adversarial embedding bank $\mathbf{A}$.
Subsequently, the similarity scores $\mathbf{S}=\mathbf{B}\mathbf{T} \in \mathbb{R}^{b \times K}$ can be calculated, with each element representing the prediction score in the following manner:
\begin{equation}\label{eq3}
    p(i, j) = p(j|e'_i) = \frac{\exp(\text{sim}(e'_i, w_j) / \tau)}{\sum_{k=1}^{K} \exp(\text{sim}(e'_{i}, w_k) / \tau)}.
\end{equation}

The learning objective during fine-tuning on the downstream dataset, aimed at maximizing the ground-truth class score, employs the cross-entropy loss function. 
Notably, at this stage, the image encoder has been discarded, and gradients are backpropagated through the text encoder to update the learnable vectors, while the text encoder is frozen.
This procedure is systematically outlined in \cref{algorithm}.

\section{Experiments}

In this section, we begin by comparing the adversarial robustness of our proposed approach with hand-crafted prompts under both white-box and black-box adversarial attacks. 
Second, we compare our method with the state-of-the-art input denoising defensive approaches.
Additionally, we investigate the trade-off between generalizability and adversarial robustness in the context of prompt tuning.
We also discuss the efficiency between our method and AT.
Next, we examine the performance of learnable vectors when trained on a specific dataset but evaluated across various distinct datasets.
Finally, we carry out an experimental analysis into interpreting the learnable vectors and perform an exhaustive analysis of hyperparameters.

\subsection{Experimental Settings}

\paragraph{Datasets.}
We conduct our study mainly on 8 high-resolution vision datasets: Pets~\cite{parkhi2012cats}, Flowers~\cite{nilsback2008automated}, ImageNet~\cite{russakovsky2015imagenet}, Food101~\cite{bossard2014food}, SUN397~\cite{xiao2010sun}, DTD~\cite{cimpoi2014describing}, EuroSAT~\cite{helber2019eurosat}, and UCF101~\cite{soomro2012ucf101}.
We adhered to the division of training and testing sets as established in the setup of \cite{zhou2022learning}.
For the ImageNet test set, in a manner consistent with prior studies focusing on adversarial attacks~\cite{dong2018boosting, wang2021enhancing, xie2019improving}, we use 1,000 images which are randomly sampled (one image per class).
Furthermore, to assess the domain generalization capabilities, we employed four variant datasets of ImageNet, namely ImageNetV2~\cite{recht2019imagenet}, ImageNet-Sketch~\cite{wang2019learning}, ImageNet-A~\cite{hendrycks2021natural}, and ImageNet-R~\cite{hendrycks2021many}.

\paragraph{Models.}
Our experiments are centered on the CLIP model. We selected the publicly available version ViT-B/16, and ViT-L/14~\cite{dosovitskiy2020image}, which has the largest parameter.
Consistent with the vanilla CLIP, we employed hand-crafted prompts as textual input, such as ``\texttt{a photo of a <class>, a type of pet}'' for Pets.

\paragraph{Adversarial Attacks.}
To evaluate adversarial robustness, we introduced both white-box and black-box adversarial attacks. 
For white-box adversarial attacks, we employed PGD-40~\cite{madry2017towards}, aimed at maximizing KL Divergence in the embedding space, and AutoAttack~\cite{croce2020reliable}, aimed at maximizing the contrastive loss between image-text pairs, respectively.
Regarding black-box attacks, we implemented black-box attack RAP~\cite{qin2022boosting}. 

\paragraph{Adversarial Defenses.}
To facilitate comparison with input denoising defenses, we incorporated two distinct categories of defense methods. One is the most effective but relatively time-consuming purification approach based on diffusion model, namely DiffPure~\cite{nie2022diffusion}. The other is a more immediate but slightly less effective method, including Super resolution~\cite{mustafa2019image} and Rescale~\cite{xie2018mitigating}.

\paragraph{Implementation Details.}
Our methodology builds upon the CoOp framework\footnote{\url{https://github.com/KaiyangZhou/CoOp}}. 
Our training process consists of 5 epochs with a batch size of 512 on ImageNet, and 100 epochs with a batch size of 32 on other datasets.
The learnable vectors are optimized via SGD, starting with an initial learning rate of 0.002 for ViT-L/14 and 0.005 for ViT-B/16, and adjusted by cosine annealing.
The number of learnable vector $M=32$.
To construct the adversarial embedding bank $\mathbf{A}$, we apply the PGD-10 attack with a maximum perturbation of 8/255 over 10 iterations.
For white-box adversarial attack on the test set, we utilize PGD-40 with a maximum perturbation of 16/255 over 40 iterations.
We conduct black-box adversarial attacks on the test set using RAP for 400 iterations.
For the selection of the RAP attack surrogate model, we employ ResNet-50 with torchvision weights for ImageNet, and train an additional fully connected layer on downstream datasets.
The hyperparameter $\sigma$ in Super-resolution was set to 0.2.
The pre-trained diffusion models in DiffPure is Guided Diffusion~\cite{dhariwal2021diffusion} and the time step was set to 150.

\subsection{Comparison with Vanilla CLIP}

\begin{table*}[t]
\centering
\caption{Accuracy (\%) under PGD-40 and RAP attacks: The ``{\textcolor{deepgreen}{( ↑)}}'' indicates the margin by which \emph{AdvPT} surpasses the vanilla CLIP (hand-crafted prompts).}
\label{tab:1}
\resizebox{1.0\linewidth}{!}{
\setlength{\tabcolsep}{1.0mm}{
\begin{tabular}{
  l 
  l 
  l llllllll
}
\toprule
\multicolumn{3}{l}{} & \textbf{Flowers} & \textbf{Pets} & \textbf{Food101} & \textbf{SUN397} & \textbf{DTD} & \textbf{EuroSAT} & \textbf{UCF101} & \textbf{ImageNet} \\
\hline
\multirow{7}{*}{\rotatebox{90}{\textbf{ViT-B/16}}} & 
\multirow{3}{*}{vanilla} & 
Clean & 71.4 & 89.1 & 86.1 & 62.6 & 44.4 & 47.8 & 66.7 & 66.1 \\
 &  & PGD & 6.4 & 24.4 & 14.0 & 14.7 & 11.1 & 22.2 & 9.1 & 6.6 \\
 &  & RAP & 60.7 & 79.9 & 68.3 & 55.4 & 33.5 & 19.2 & 56.4 & 28.6 \\
\cline{2-11}
 & \multirow{3}{*}{\emph{AdvPT}} & 
Clean & 87.6 & 91.3 & 84.4 & 70.7 & 67.9 & 68.1 & 77.0 & 69.1 \\
 &  & PGD & 37.4\small {\color{deepgreen}{(31.0↑)}}  & 41.9\small {\color{deepgreen}{(17.5↑)}} & 38.8\small{\color{deepgreen}{(24.8↑)}} & 35.7\small {\color{deepgreen}{(21.0↑)}} & 39.7\small {\color{deepgreen}{(28.6↑)}} & 55.4\small {\color{deepgreen}{(33.2↑)}} & 27.2\small {\color{deepgreen}{(18.1↑)}} & 19.9\small {\color{deepgreen}{(13.3↑)}} \\
 &  & RAP & 79.0\small {\color{deepgreen}{(18.3↑)}} & 81.8\small {\color{deepgreen}{(1.9↑)}} & 68.7\small {\color{deepgreen}{(0.4↑)}} & 60.0\small {\color{deepgreen}{(4.6↑)}} & 50.5\small {\color{deepgreen}{(17.0↑)}} & 40.6\small {\color{deepgreen}{(21.4↑)}} & 66.0\small {\color{deepgreen}{(9.6↑)}} & 30.2\small {\color{deepgreen}{(1.6↑)}} \\
\midrule
\multirow{7}{*}{\rotatebox{90}{\textbf{ViT-L/14}}} & 
\multirow{3}{*}{vanilla} & 
Clean & 79.3 & 93.6 & 91.0 & 67.6 & 53.1 & 58.1 & 74.2 & 72.8 \\
 &  &  PGD & 20.1 & 50.3 & 34.3 & 27.9 & 20.7 & 23.3 & 33.9 & 28.5 \\
 &  & RAP & 70.6 & 88.2 & 81.9 & 62.5 & 42.5 & 42.3 & 67.3 & 40.2 \\
\cline{2-11}
 & \multirow{3}{*}{\emph{AdvPT}} & 
Clean & 97.6 & 92.9 & 90.9 & 76.4 & 72.8 & 79.2 & 86.5 & 77.8 \\
 &  & PGD & 56.0\small {\color{deepgreen}{(35.9↑)}} & 68.7\small {\color{deepgreen}{(18.4↑)}} & 54.0\small {\color{deepgreen}{(19.7↑)}} & 44.0\small {\color{deepgreen}{(16.1↑)}} & 42.0\small {\color{deepgreen}{(21.3↑)}} & 62.2\small {\color{deepgreen}{(38.9↑)}} & 47.9\small {\color{deepgreen}{(14.0↑)}} & 42.9\small {\color{deepgreen}{(14.4↑)}} \\
 &  & RAP & 94.1\small {\color{deepgreen}{(23.5↑)}} & 90.4\small {\color{deepgreen}{(2.2↑)}} & 82.7\small {\color{deepgreen}{(0.8↑)}} & 70.3\small {\color{deepgreen}{(7.8↑)}} & 62.4\small {\color{deepgreen}{(19.9↑)}} & 50.8\small {\color{deepgreen}{(8.5↑)}} & 78.7\small {\color{deepgreen}{(11.4↑)}} & 47.6\small {\color{deepgreen}{(7.4↑)}} \\
\bottomrule
\end{tabular}}}
\end{table*}

We started our evaluation by comparing \emph{AdvPT} with the vanilla CLIP model. Using PGD-40 and RAP, we evaluated adversarial robustness in 8 datasets, as indicated in \cref{tab:1}. Our findings reveal that: 
(1) \emph{AdvPT} demonstrates improvements over the vanilla CLIP under both PGD-40 and RAP attacks, with the specific improvement quantified in \textcolor{deepgreen}{green}.
(2) While the primary goal of \emph{AdvPT} is not to enhance generalizability, the empirical finding implies that the enhancement of accuracy emerges as a collateral advantage.

\subsection{Comparison with Adversarial Defenses}

\begin{table*}[t]
\centering
\caption{Accuracy (\%) under PGD-40 Attack: The ``\colorbox{gray!25}{\textcolor{deepgreen}{\emph{+AdvPT}}}'' indicates our method combined with the input denoising defense. The best results are highlighted in \textbf{bold}.}
\label{tab:2}
\resizebox{1.0\linewidth}{!}{
\setlength{\tabcolsep}{1.2mm}{
\begin{tabular}{
  l 
  l 
  l
  l
  l
  l
  l
  l
  l
  l
}
\toprule
\multicolumn{2}{l}{} & \textbf{Flowers} & \textbf{Pets} & \textbf{Food101} & \textbf{SUN397} & \textbf{DTD} & \textbf{EuroSAT} & \textbf{UCF101} & \textbf{ImageNet} \\
\hline
 & No   defense & 6.4 & 24.4 & 14.0 & 14.7 & 11.1 & 22.2 & 9.1 & 6.6\\
 \rowcolor{gray!25} \cellcolor{white}& \emph{AdvPT} & 37.4\small {\color{deepgreen}{(31.0↑)}} & 41.9\small {\color{deepgreen}{(17.5↑)}} & 38.8\small {\color{deepgreen}{(24.8↑)}} & 35.7\small {\color{deepgreen}{(21.0↑)}} & 39.7\small {\color{deepgreen}{(28.6↑)}} & 55.4\small {\color{deepgreen}{(33.2↑)}} & 27.2\small {\color{deepgreen}{(18.1↑)}} & 19.9\small {\color{deepgreen}{(13.3↑)}} \\

 & Super & 13.8 & 43.6 & 58.1 & 40.5 & 32.1 & 43.3 & 35.4 & 18.3 \\
\rowcolor{gray!25} \cellcolor{white}& \textcolor{deepgreen}{\emph{+AdvPT}} & 60.4\small {\color{deepgreen}{(46.6↑)}} & 68.3\small {\color{deepgreen}{(24.7↑)}}& 69.9\small {\color{deepgreen}{(11.8↑)}} & 69.9\small {\color{deepgreen}{(29.4↑)}}& 58.2\small {\color{deepgreen}{(26.1↑)}} & \textbf{76.7}\small {\color{deepgreen}{(33.4↑)}} & 58.4\small {\color{deepgreen}{(23.0↑)}}& 34.9\small {\color{deepgreen}{(16.6↑)}}\\

 & DiffPure & 59.4 & 84.1 & 68.6 & 55.0 & 36.9& 29.7 & 60.4 & 56.6 \\
\rowcolor{gray!25} \cellcolor{white}& \textcolor{deepgreen}{\emph{+AdvPT}} & 81.9\small {\color{deepgreen}{(22.5↑)}}  & 86.9\small {\color{deepgreen}{(2.8↑)}}  & 70.5\small {\color{deepgreen}{(1.9↑)}} & 63.9\small {\color{deepgreen}{(8.9↑)}} & 60.8\small {\color{deepgreen}{(23.9↑)}}  & 59.6\small {\color{deepgreen}{(29.9↑)}}  & \textbf{72.2}\small {\color{deepgreen}{(11.8↑)}}  & 61.1\small {\color{deepgreen}{(4.5↑)}} \\

 & Rescale & 60.1 & 81.9 & 79.0 & 56.9 & 39.9 & 40.6 & 58.6 & 53.3 \\
\rowcolor{gray!25} \cellcolor{white} \multirow{-8}{*}{\rotatebox{90}{\textbf{ViT-B/16}}} & \textcolor{deepgreen}{\emph{+AdvPT}} & \textbf{87.5}\small {\color{deepgreen}{(27.4↑)}}& \textbf{87.4}\small {\color{deepgreen}{(5.5↑)}}& \textbf{80.4}\small {\color{deepgreen}{(1.4↑)}}& \textbf{67.1}\small {\color{deepgreen}{(10.2↑)}} & \textbf{64.4}\small {\color{deepgreen}{(24.5↑)}} & 75.4\small {\color{deepgreen}{(34.8↑)}} & 72.1\small {\color{deepgreen}{(13.5↑)}}& \textbf{61.6}\small {\color{deepgreen}{(8.3↑)}}\\
\midrule
& No   defense & 20.1 & 50.3 & 34.3 & 27.9 & 20.7 & 23.3 & 33.9 & 28.5 \\
 \rowcolor{gray!25} \cellcolor{white}& \emph{AdvPT} & 56.0\small {\color{deepgreen}{(35.9↑)}}& 68.7\small {\color{deepgreen}{(18.4↑)}}& 54.0\small {\color{deepgreen}{(19.7↑)}} & 44.0\small {\color{deepgreen}{(16.1↑)}}& 42.0\small {\color{deepgreen}{(21.3↑)}}& 62.2\small {\color{deepgreen}{(38.9↑)}} & 47.9\small {\color{deepgreen}{(14.0↑)}}& 42.9\small {\color{deepgreen}{(14.5↑)}} \\

 & Super & 31.6 & 67.7 & 51.0 & 39.5 & 34.5 & 45.3 & 52.7 &  40.3\\
\rowcolor{gray!25} \cellcolor{white}& \textcolor{deepgreen}{\emph{+AdvPT}} & 74.9\small {\color{deepgreen}{(43.3↑)}} & 81.2\small{\color{deepgreen}{(13.5↑)}}  & 68.2\small{\color{deepgreen}{(17.2↑)}}  & 55.5\small{\color{deepgreen}{(16.0↑)}} & 59.0\small{\color{deepgreen}{(24.5↑)}}  & 80.1\small{\color{deepgreen}{(34.8↑)}} & 70.3\small{\color{deepgreen}{(17.6↑)}}  & 54.3\small{\color{deepgreen}{(14.0↑)}} \\

 & DiffPure & 69.2 & 90.6 & 73.8 & 60.6 & 46.6 & 35.6 & 67.1 & 64.5 \\
\rowcolor{gray!25} \cellcolor{white}& \textcolor{deepgreen}{\emph{+AdvPT}} & 92.0\small{\color{deepgreen}{(22.8↑)}}  & 90.3\small{\color{red}{(0.3↓)}}  & 77.2\small{\color{deepgreen}{(3.4↑)}}  & 70.5\small{\color{deepgreen}{(9.9↑)}} & 67.3\small{\color{deepgreen}{(20.7↑)}} & 64.5\small{\color{deepgreen}{(28.9↑)}} & 79.1\small{\color{deepgreen}{(12.0↑)}} & 69.7\small{\color{deepgreen}{(5.2↑)}}\\

 & Rescale & 73.2 & 88.9 & 83.0 & 63.0 & 46.7 & 46.9 & 70.2 & 66.1 \\
\rowcolor{gray!25} \cellcolor{white} \multirow{-8}{*}{\rotatebox{90}{\textbf{ViT-L/14}}} & \textcolor{deepgreen}{\emph{+AdvPT}} & \textbf{94.5}\small{\color{deepgreen}{(21.3↑)}} & \textbf{91.0}\small{\color{deepgreen}{(2.1↑)}}& \textbf{86.0}\small{\color{deepgreen}{(3.0↑)}} & \textbf{73.2}\small{\color{deepgreen}{(10.2↑)}}& \textbf{69.8}\small{\color{deepgreen}{(23.1↑)}} & \textbf{83.3}\small{\color{deepgreen}{(36.4↑)}}& \textbf{82.2}\small{\color{deepgreen}{(12.0↑)}}& \textbf{74.8}\small{\color{deepgreen}{(8.7↑)}}\\
\bottomrule
\end{tabular}}}
\end{table*}

\begin{table}[t]
\centering
\caption{Accuracy (\%) under AutoAttack.}
\label{tab:A2}
\resizebox{1.0\linewidth}{!}{
\setlength{\tabcolsep}{1.2mm}{
\begin{tabular}{
  l 
  l 
  l
  l
  l
  l
  l
  l
  l
  l
}
\toprule
\multicolumn{2}{l}{} & \textbf{Flowers} & \textbf{Pets} & \textbf{Food101} & \textbf{SUN397} & \textbf{DTD} & \textbf{EuroSAT} & \textbf{UCF101} & \textbf{ImageNet} \\
\hline
 & No   defense &0.0 &0.0 & 0.0& 0.0 &0.0 & 0.0 &0.0 &0.0\\
 \rowcolor{gray!25} \cellcolor{white}& \emph{AdvPT} &23.3\small{\color{deepgreen}{(23.3↑)}} &7.2\small{\color{deepgreen}{(7.2↑)}} & 4.4\small{\color{deepgreen}{(4.4↑)}}& 17.6\small{\color{deepgreen}{(17.6↑)}} &28.9\small{\color{deepgreen}{(28.9↑)}} & 27.5\small{\color{deepgreen}{(27.5↑)}} &18.5\small{\color{deepgreen}{(18.5↑)}} &11.0\small{\color{deepgreen}{(11.0↑)}}\\


 & DiffPure & 54.1 & 78.9 & 61.1 & 51.5 & 35.1 & 32.9 & 56.7 & 55.5 \\
\rowcolor{gray!25} \cellcolor{white} \multirow{-4}{*}{\rotatebox{90}{\textbf{ViT-B/16}}}& \textcolor{deepgreen}{\emph{+AdvPT}} & \textbf{80.3}\small{\color{deepgreen}{(26.2↑)}}  & \textbf{84.7}\small{\color{deepgreen}{(5.8↑)}} & \textbf{65.5}\small{\color{deepgreen}{(4.4↑)}} & \textbf{61.4}\small{\color{deepgreen}{(9.9↑)}} & \textbf{60.8}\small{\color{deepgreen}{(25.7↑)}} & \textbf{59.1}\small{\color{deepgreen}{(26.2↑)}} & \textbf{70.8}\small{\color{deepgreen}{(14.1↑)}} &\textbf{60.4}\small{\color{deepgreen}{(4.9↑)}} \\

\midrule
& No   defense &0.0 &0.0 & 0.0& 0.0 &0.0 & 0.0 &0.0 &0.0\\
 \rowcolor{gray!25} \cellcolor{white}& \emph{AdvPT} & 19.2\small{\color{deepgreen}{(19.2↑)}} & 3.3\small{\color{deepgreen}{(3.3↑)}} & 3.3\small{\color{deepgreen}{(3.3↑)}} & 15.5\small{\color{deepgreen}{(15.5↑)}} & 25.7\small{\color{deepgreen}{(25.7↑)}} & 29.5\small{\color{deepgreen}{(29.5↑)}} & 17.0\small{\color{deepgreen}{(17.0↑)}} & 9.0\small{\color{deepgreen}{(9.0↑)}}\\

 & DiffPure & 63.7 & 87.5 & 67.4 & 51.5 & 43.4 & 34.8 & 64.7 & 60.7 \\
\rowcolor{gray!25} \cellcolor{white} \multirow{-4}{*}{\rotatebox{90}{\textbf{ViT-L/14}}}& \textcolor{deepgreen}{\emph{+AdvPT}} & \textbf{89.3}\small{\color{deepgreen}{(25.6↑)}} & \textbf{87.9}\small{\color{deepgreen}{(0.4↑)}} & \textbf{72.5}\small{\color{deepgreen}{(5.1↑)}} & \textbf{68.5}\small{\color{deepgreen}{(17.0↑)}}& \textbf{65.4}\small{\color{deepgreen}{(22.0↑)}} & \textbf{63.2}\small{\color{deepgreen}{(28.4↑)}} & \textbf{77.8}\small{\color{deepgreen}{(13.1↑)}} & \textbf{67.9}\small{\color{deepgreen}{(7.2↑)}}\\

\bottomrule
\end{tabular}}}
\end{table}

As described previously, \emph{AdvPT} presents an innovative approach to enhance the robustness of image encoders against adversarial attacks by modifying only the textual input.
This method is inherently synergistic with visual-modality input denoising defenses.
We evaluated its performance against white-box PGD-40, and observed its compatibility with defenses, as delineated in \cref{tab:2}.
Significantly, incorporation of \emph{AdvPT} requires no specialized tuning for the purified images.

Our results show \emph{AdvPT}'s consistent compatibility with benchmark adversarial defenses. 
Despite a minor 0.3\% performance drop in ViT-L14 on Pets, it maintains over 90\% accuracy, closely paralleling original example performance, which is acceptable.
All improvements are highlighted in \textcolor{deepgreen}{green}, corroborating the efficacy of the strategy that combines \emph{AdvPT} with input denoising mechanisms.

Remarkably, the synergy of \emph{AdvPT} with baseline defense mechanisms sometimes yielded ``$1+1>2$'' contribution.
For example, on the ViT-B/16 model applied to the Flowers dataset, \emph{AdvPT} alone increases accuracy by 31.0\% (from 6.40\% to 37.40\%), yet when combined with Super-resolution, it further improves the performance of Super-resolution by 46.60\% (from 13.80\% to 60.4\%).
In addition to this, we also introduced AutoAttack, which targets the contrastive loss of image-text pairs. The results in \cref{tab:A2} are consistent with those in \cref{tab:2}, indicating that when combined with the state-of-the-art diffusion model-based defense method, DiffPure, our \emph{AdvPT} achieved enhanced performance.
These findings highlight the potential of this innovative synergy strategy to enhance adversarial defense by simultaneously modifying textual and visual inputs, warranting further investigation in future study.

\subsection{Generalization-Robustness Trade-off}\label{sec:5-4}

\begin{figure*}[t]
  \centering
  \begin{minipage}{0.24\textwidth}
    \includegraphics[width=\linewidth]{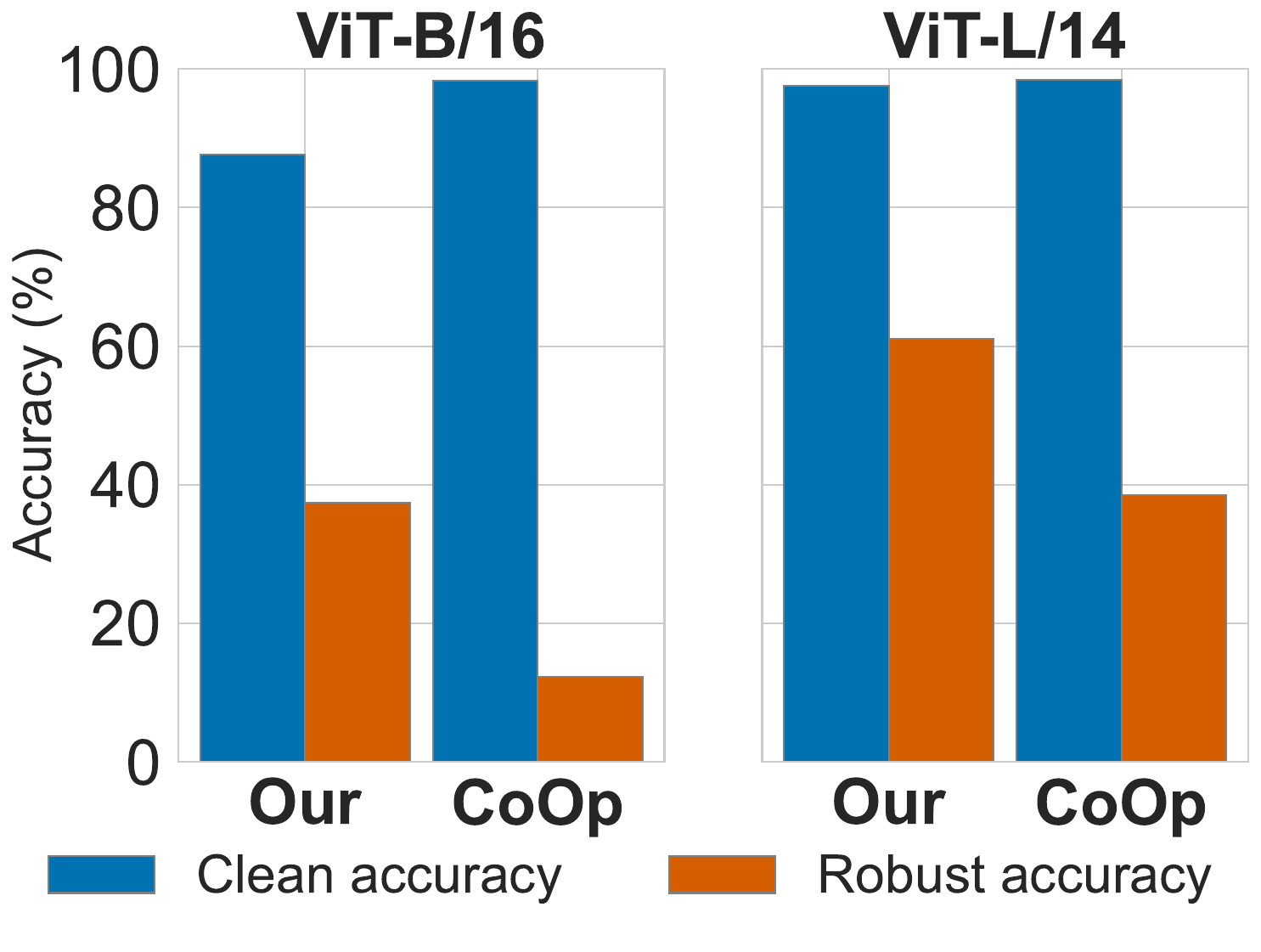} 
    \centerline{(a) Flowers}
  \end{minipage}\hfill
  \begin{minipage}{0.24\textwidth}
    \includegraphics[width=\linewidth]{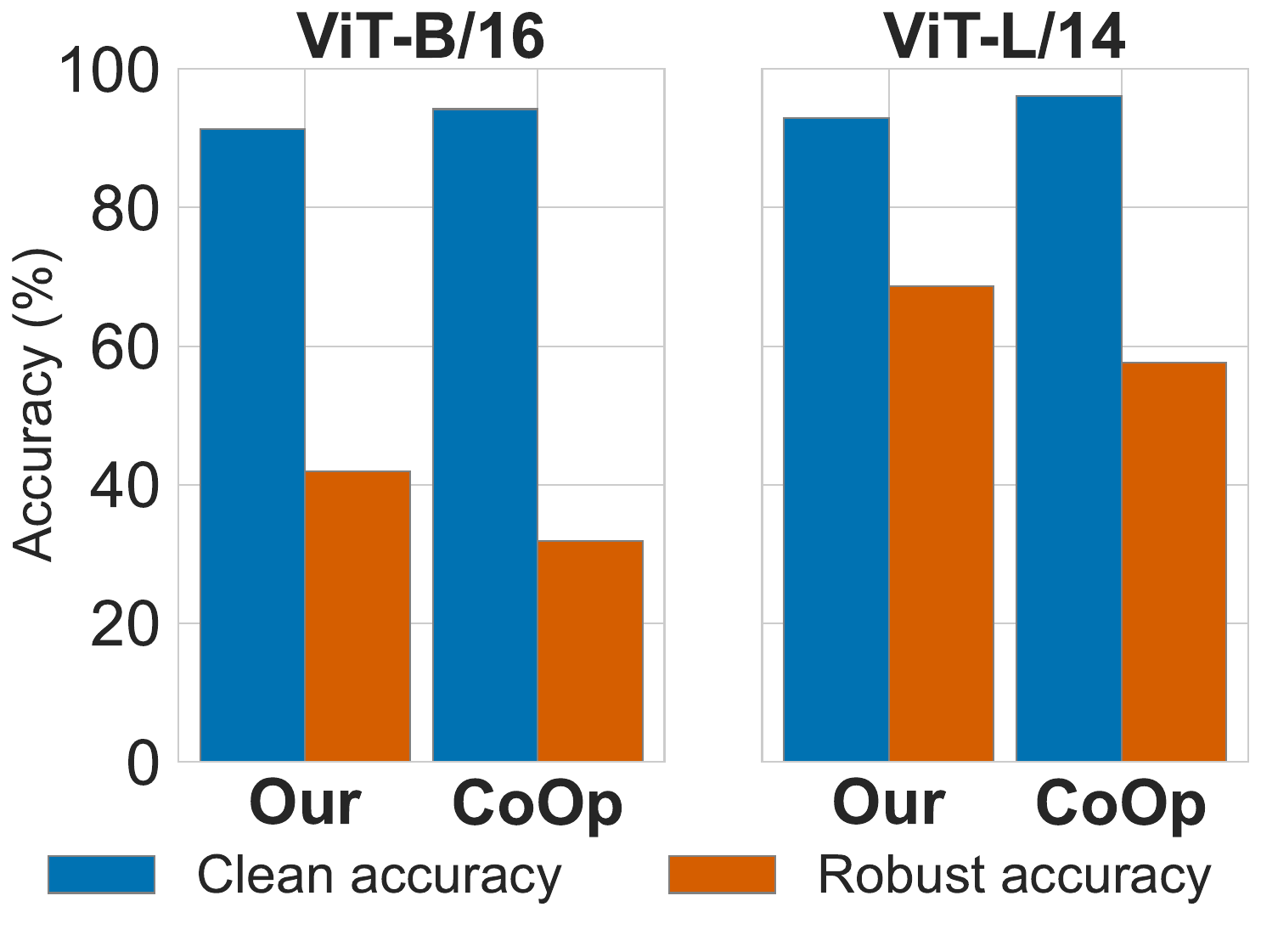}
    \centerline{(b) Pets}
  \end{minipage}\hfill
  \begin{minipage}{0.24\textwidth}
    \includegraphics[width=\linewidth]{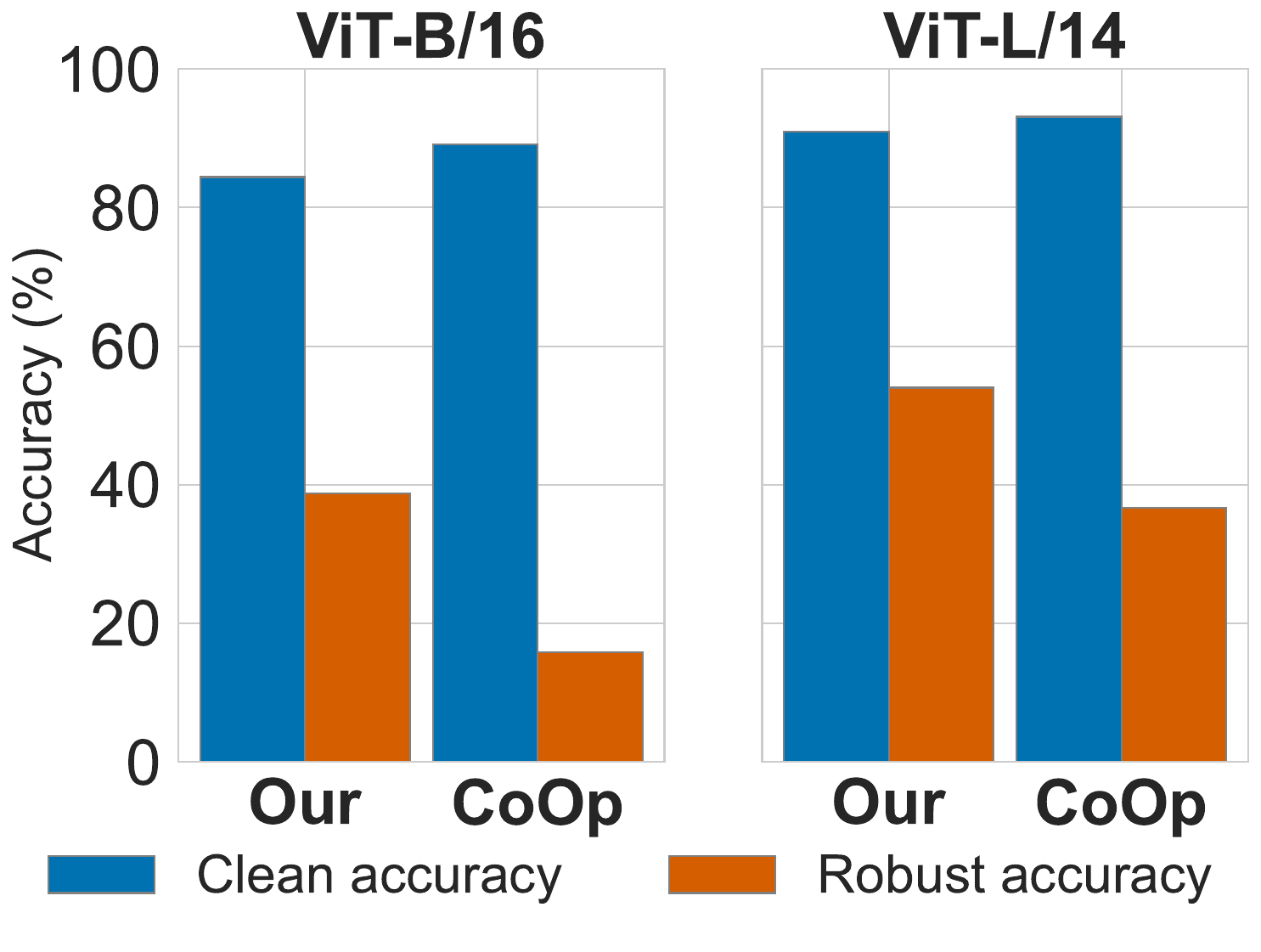}
    \centerline{(c) Food101}
  \end{minipage}\hfill
  \begin{minipage}{0.24\textwidth}
    \includegraphics[width=\linewidth]{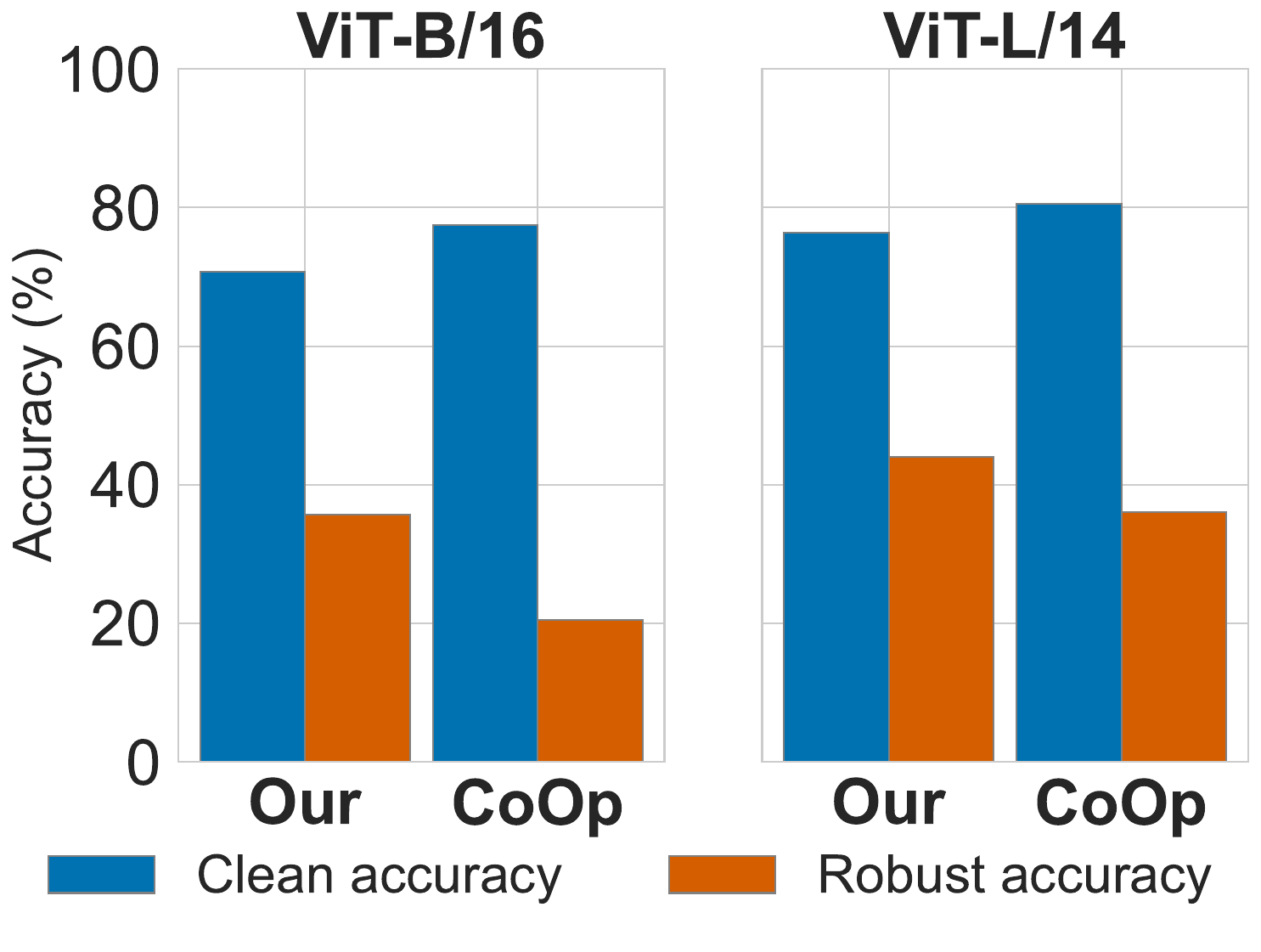}
    \centerline{(d) SUN397}
  \end{minipage}

  \begin{minipage}{0.24\textwidth}
    \includegraphics[width=\linewidth]{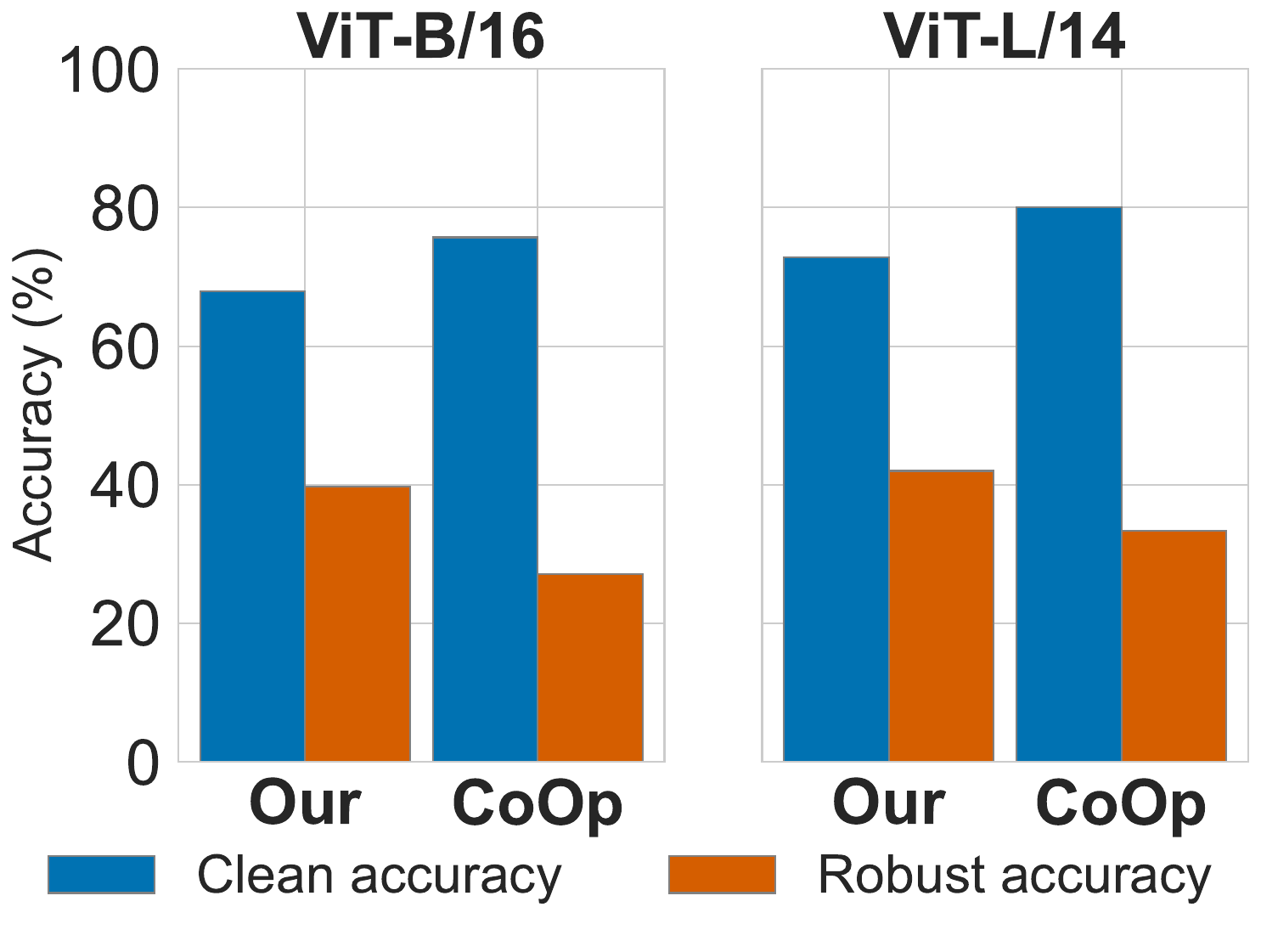}
    \centerline{(e) DTD}
  \end{minipage}\hfill
  \begin{minipage}{0.24\textwidth}
    \includegraphics[width=\linewidth]{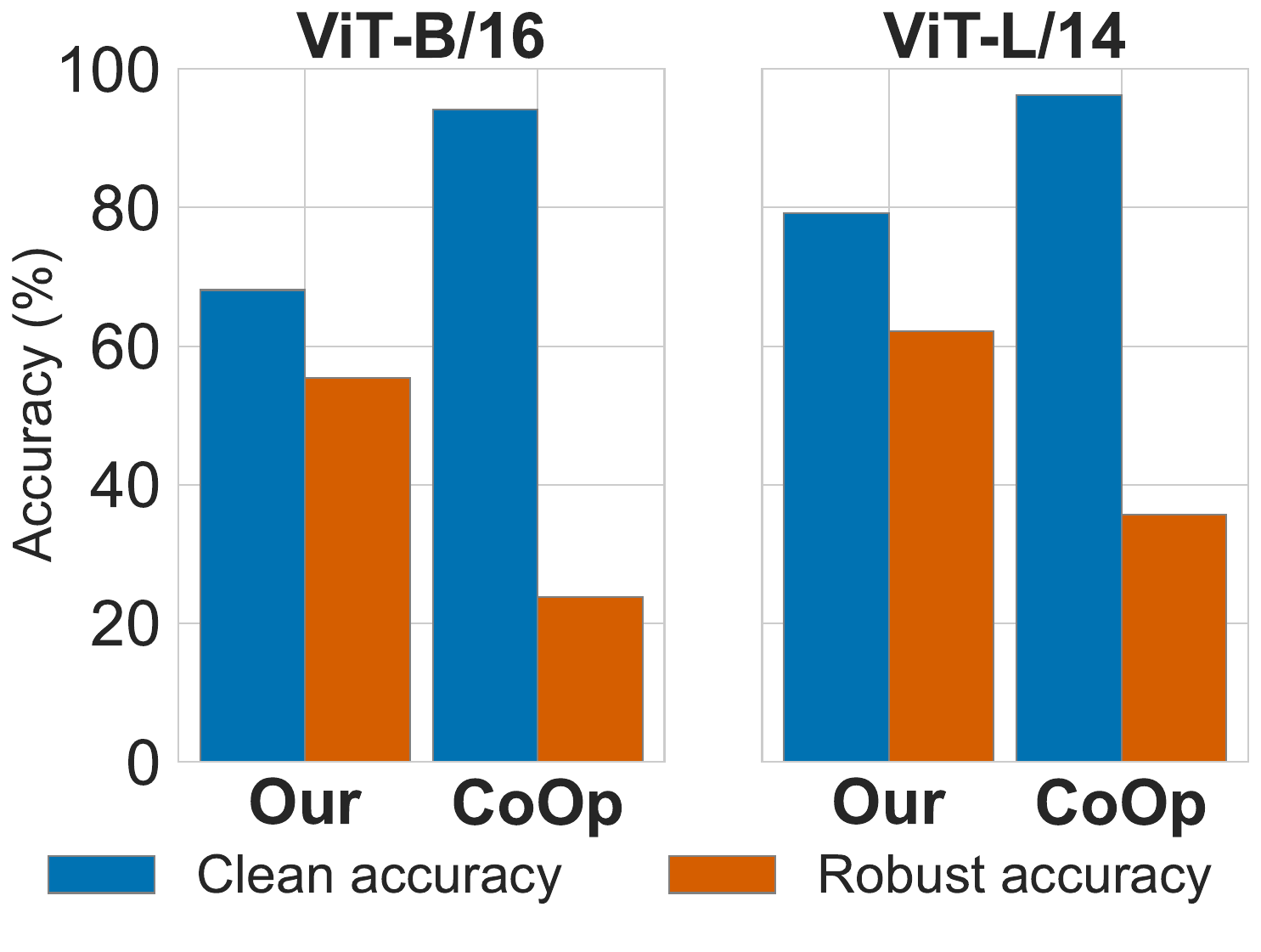}
    \centerline{(f) EuroSAT}
  \end{minipage}\hfill
  \begin{minipage}{0.24\textwidth}
    \includegraphics[width=\linewidth]{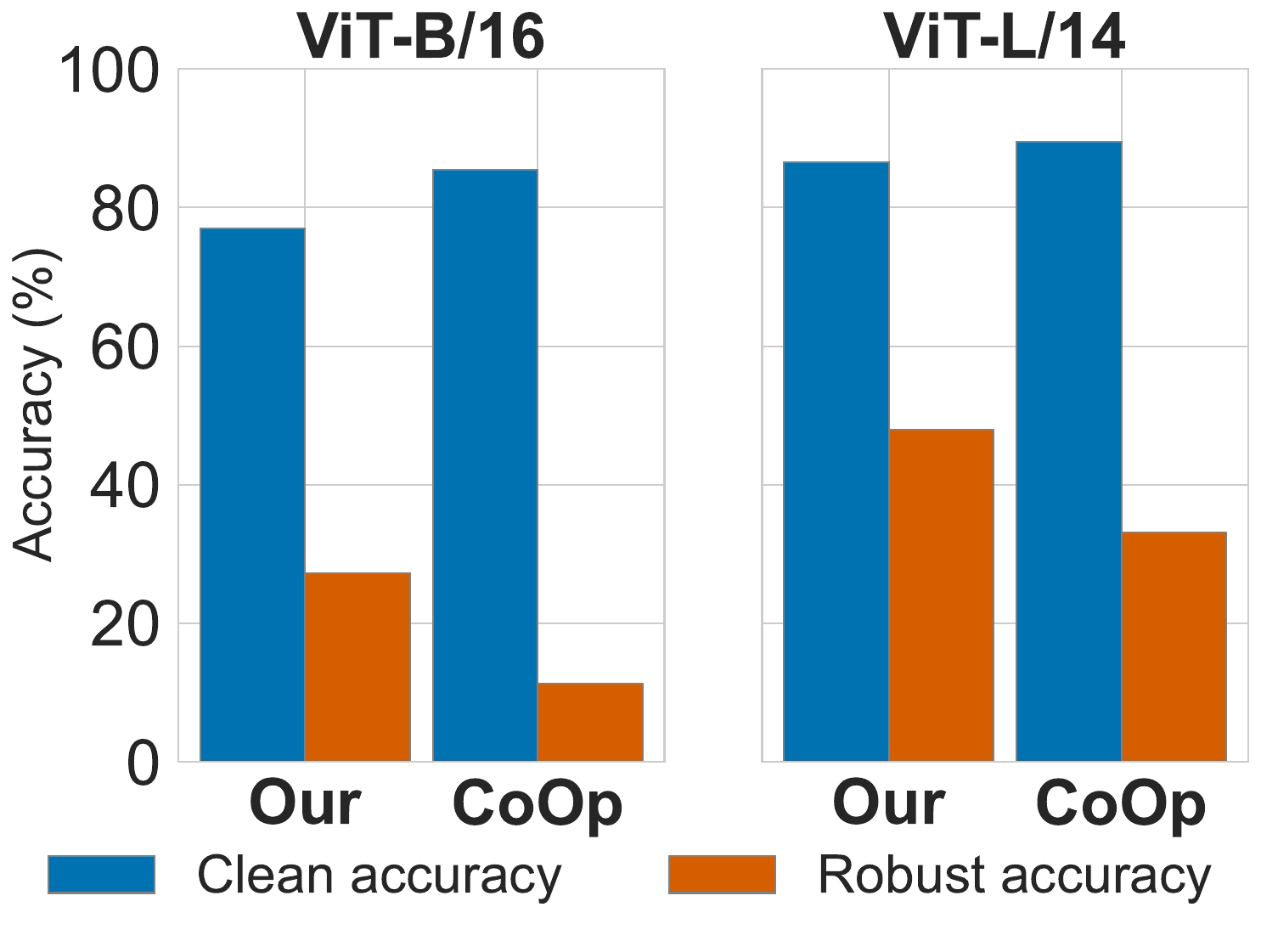}
    \centerline{(g) UCF101}
  \end{minipage}\hfill
  \begin{minipage}{0.24\textwidth}
    \includegraphics[width=\linewidth]{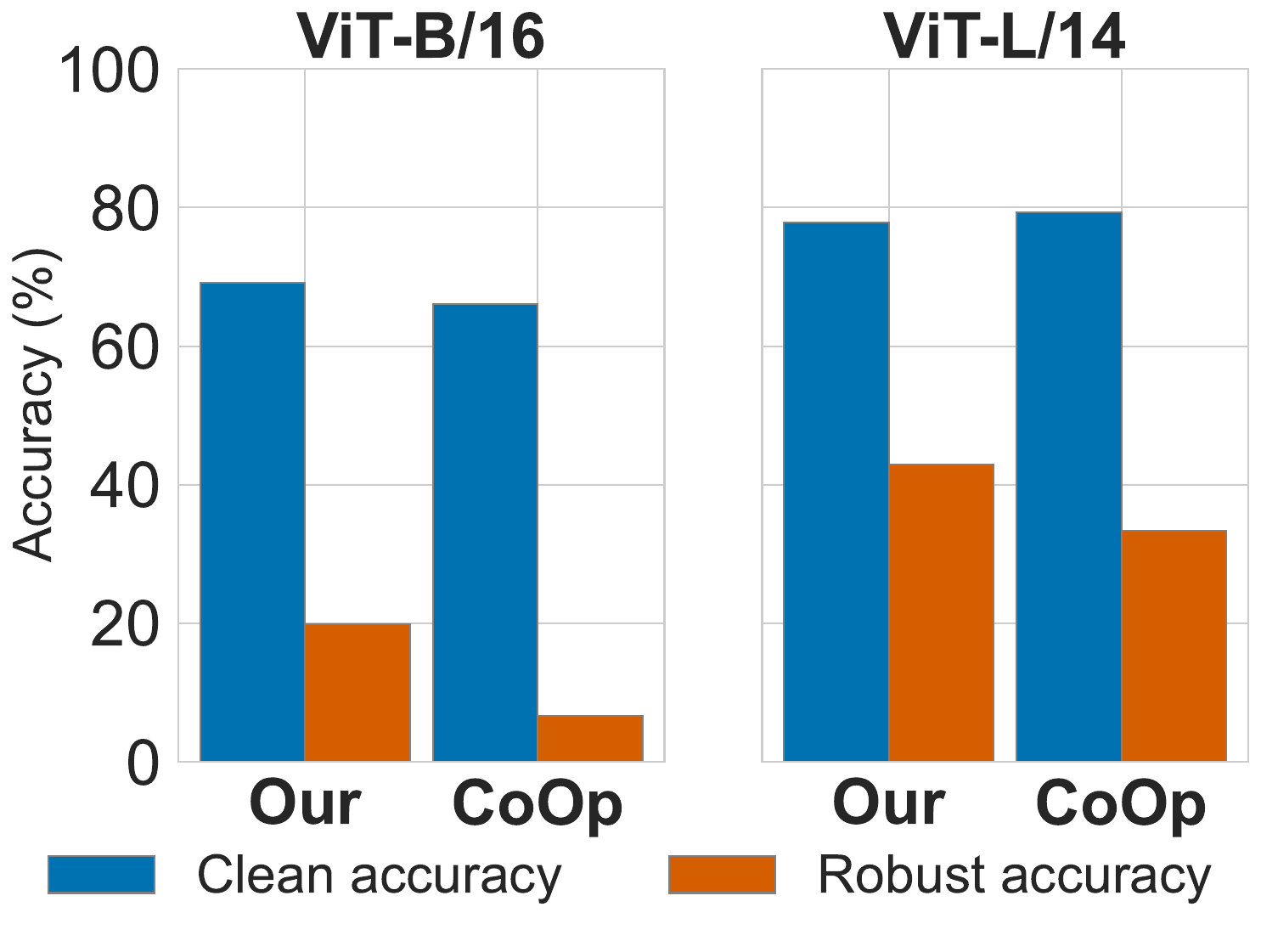}
    \centerline{(h) ImageNet}
  \end{minipage}
  
  \caption{\emph{AdvPT} vs. CoOp on generalization and adversarial robustness.}\label{fig:3}
\end{figure*}

In this subsection, we discuss the effects of various learning objectives on the learnable vectors within prompt tuning. 
The primary goal of \emph{AdvPT} is to enhance the adversarial robustness of the image modality in VLMs. 
In contrast, we explore whether an objective like CoOp~\cite{zhou2022learning}, which is fine-tuned on clean images for improved accuracy, affects adversarial robustness differently.

Our comparative analysis of \emph{AdvPT} and CoOp unveils insights into their generalizability and adversarial robustness, as illustrated in \cref{fig:3}.
Our findings are twofold: (1) Intriguingly, \emph{AdvPT} significantly outperforms CoOp in adversarial robustness, albeit at a slight cost to generalization.
This highlights a potential trade-off between adversarial robustness and generalization in prompt tuning, aligning with conclusions drawn from traditional AT~\cite{zhang2019theoretically}.
(2) Although \emph{AdvPT} sacrifices some generalizability, this drawback is mitigated as the model scale increases.
Particularly on ViT-L/14, while also enhancing adversarial robustness, the narrowed generalizability gap makes \emph{AdvPT} highly compatible with the ongoing trend towards scaling up models.

\subsection{Comparison with Adversarial Training}

\begin{figure*}[t]
  \centering
  \begin{minipage}{0.49\linewidth}
    \includegraphics[width=\linewidth]{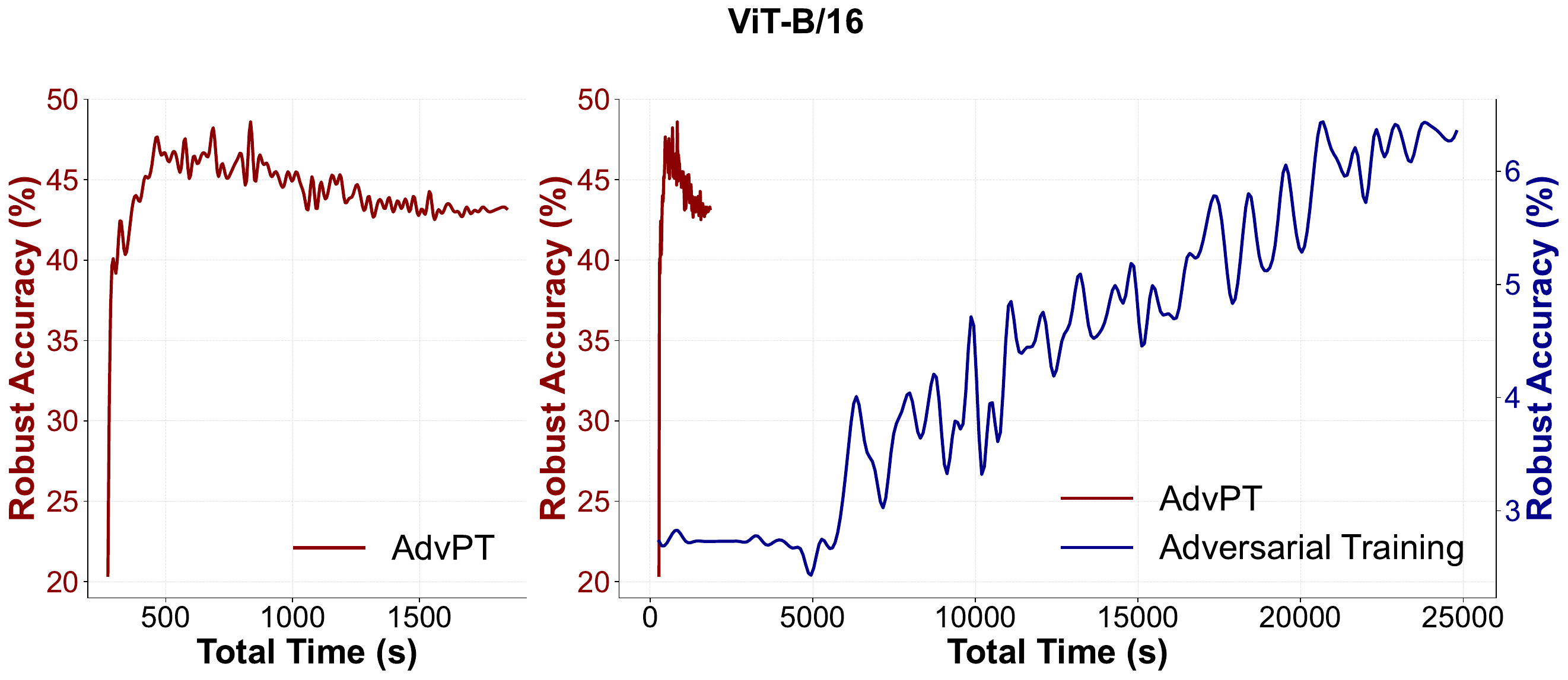}
  \end{minipage}\hfill
    \begin{minipage}{0.49\linewidth}
    \includegraphics[width=\linewidth]{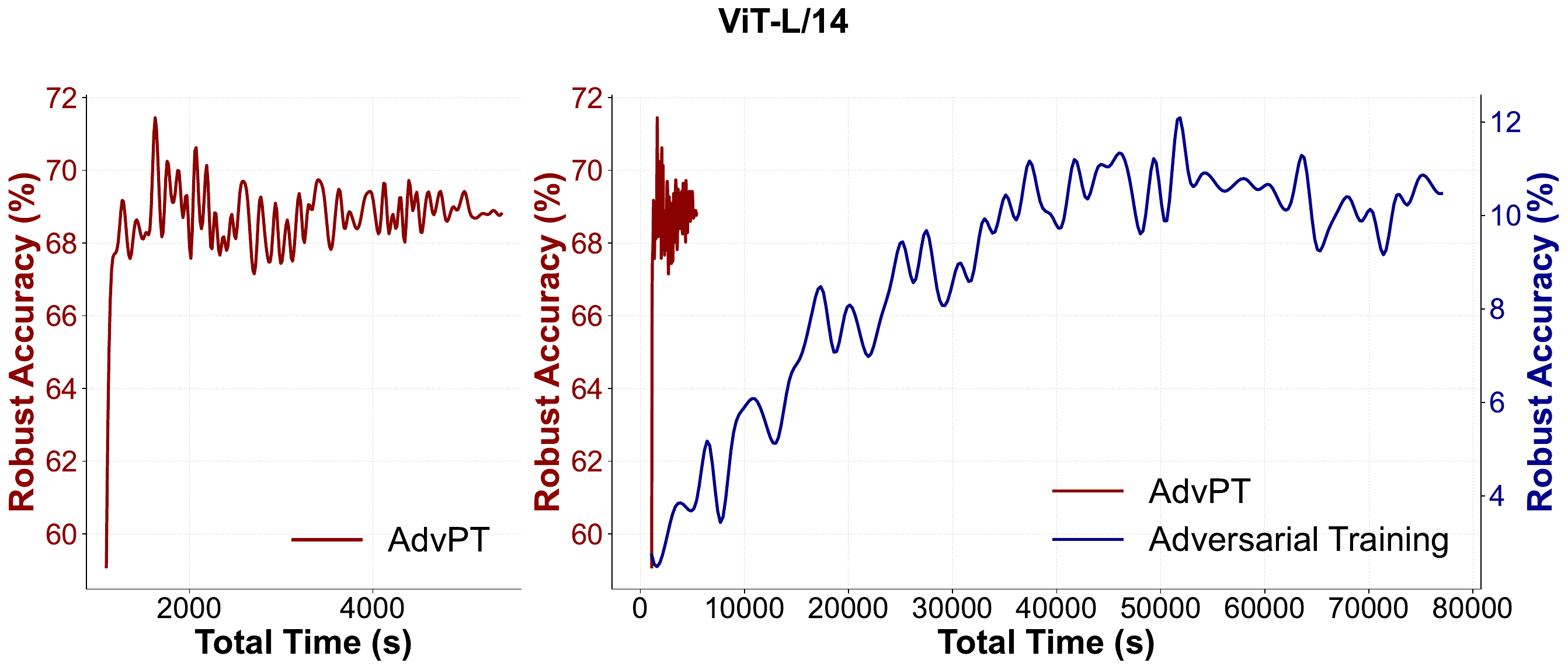}
  \end{minipage}
  \caption{Efficiency comparison between \emph{AdvPT} and AT on Pets.}
   \label{fig:5}
\end{figure*}

\begin{wrapfigure}{r}{0.6\textwidth}
\begin{minipage}[b]{0.49\linewidth}
\centering
\includegraphics[width=\textwidth]{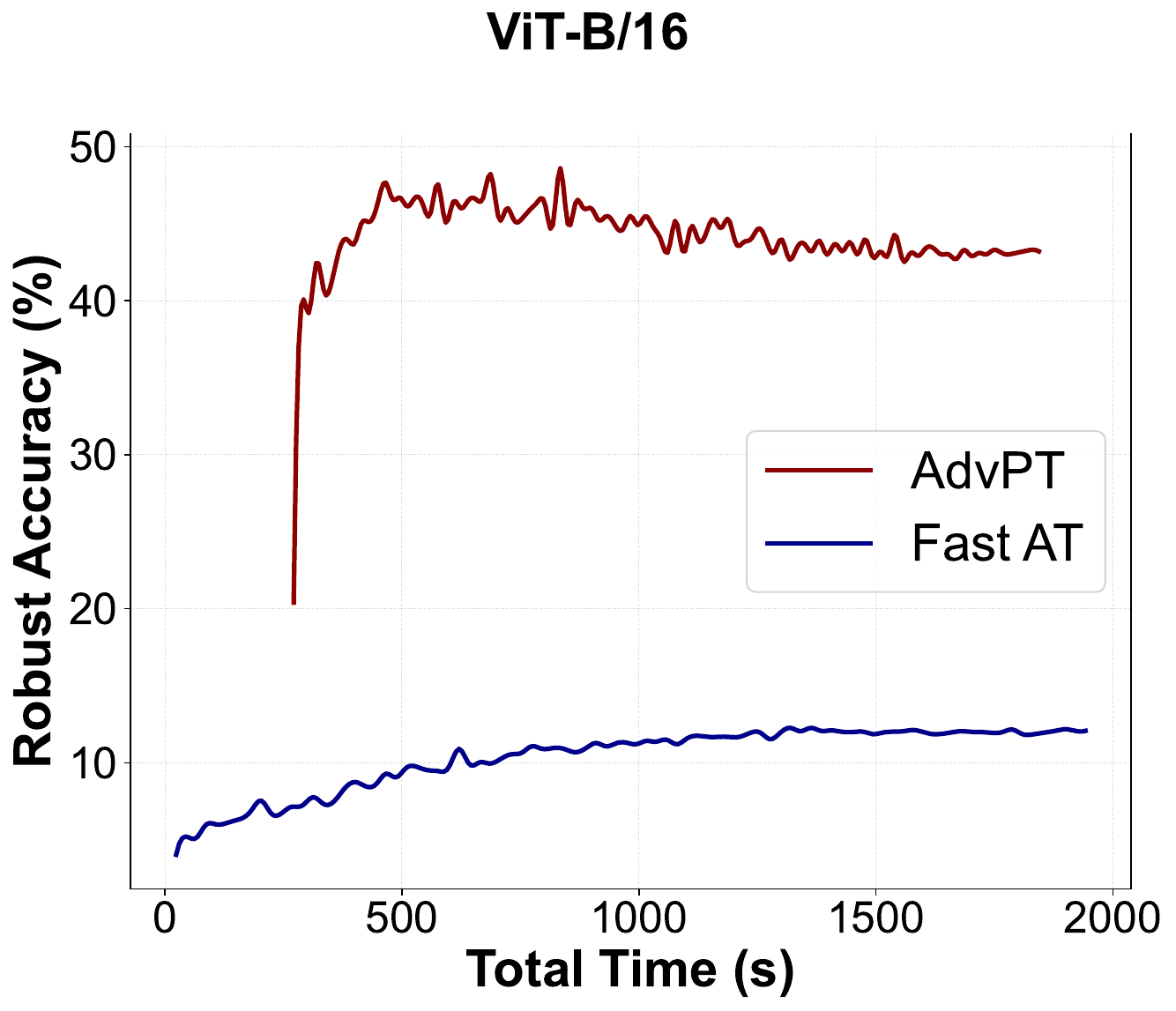}
\end{minipage}
\begin{minipage}[b]{0.49\linewidth}
\centering
\includegraphics[width=\textwidth]{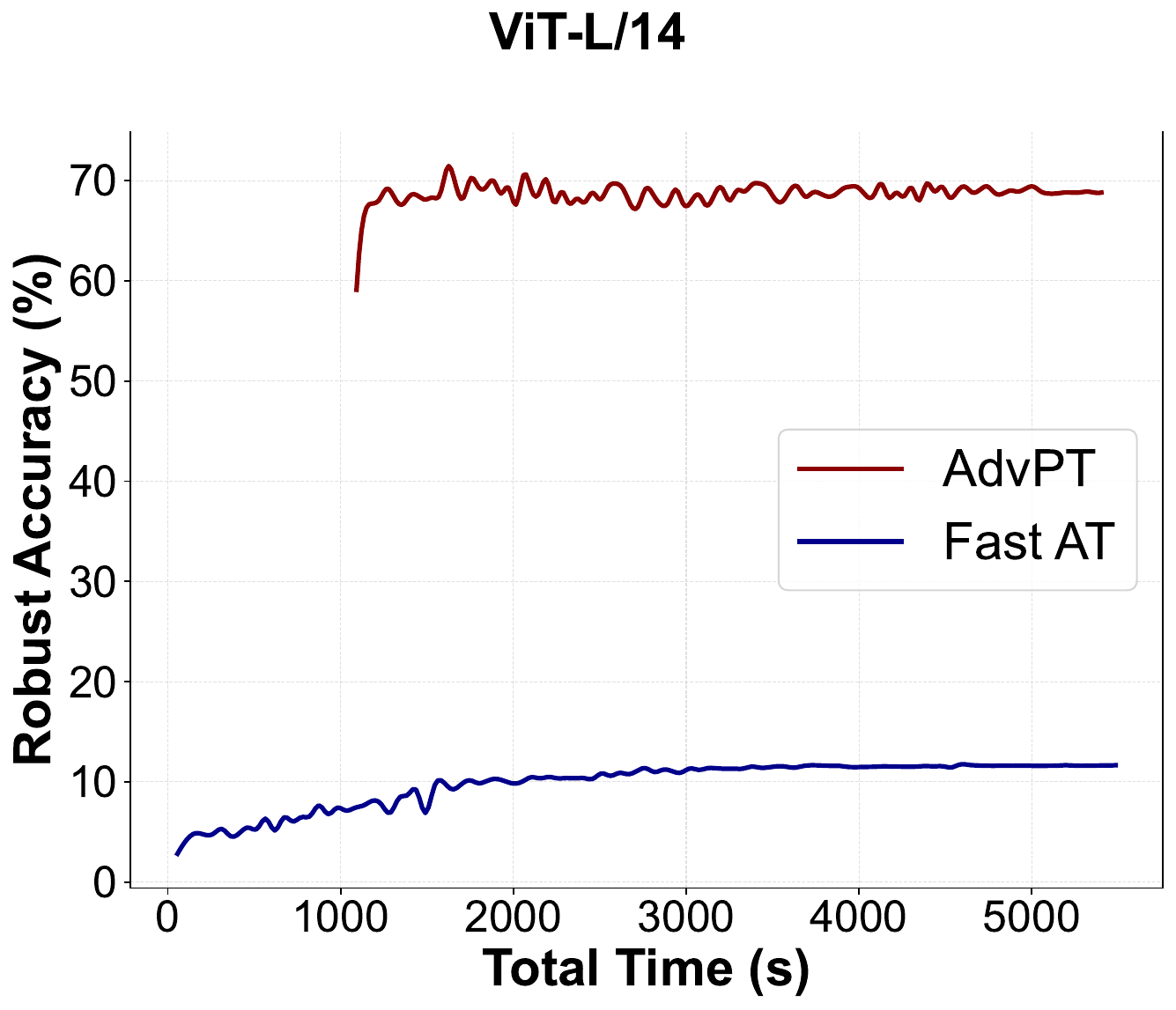}
\end{minipage}
\caption{Efficiency comparison between \emph{AdvPT} and Fast AT on Pets.}
 \label{figA:5}
\end{wrapfigure}

In this subsection, we compare the efficiency of our method, which focuses on fine-tuning only the prompt, against traditional AT. 
Specifically, we juxtapose \emph{AdvPT} with PGD-10 AT~\cite{madry2017towards} on the Pets dataset, ensuring that both methods use an equivalent batch size. 
The comparative results are shown in \cref{fig:5}, including the time taken to compute the adversarial embedding bank $\mathbf{A}$ in the total time reported.
We also presented the results of Fast AT~\cite{wongfast2020} in \cref{figA:5}. 
Although Fast AT is much faster than AT, it still lags significantly behind \emph{AdvPT}.

Our analysis reveals that \emph{AdvPT} is more time-efficient than AT, requiring at least an order of magnitude less time. 
Moreover, it demonstrates a superior enhancement in model performance, outperforming AT by at least one order of magnitude in effectiveness. 
This efficiency advantage, exceeding 100$\times$ at least, positions \emph{AdvPT} as a notably superior solution for VLMs.

\subsection{Comparison with Linear Prob CLIP}

\begin{table}[t]
\centering
\caption{Clean accuracy and robust accuracy (PGD-40) of linear prob CLIP.}
\label{tab:A1}
\resizebox{1.0\linewidth}{!}{
\setlength{\tabcolsep}{2.0mm}{
\begin{tabular}{
  l 
  l 
  c ccccccc
}
\toprule
 & &\textbf{Flowers} & \textbf{Pets} & \textbf{Food101} & \textbf{SUN397} & \textbf{DTD} & \textbf{EuroSAT} & \textbf{UCF101}  \\
\hline
\multirow{2}{*}{\rotatebox{0}{\textbf{ViT-B/16}}} & 
Clean & 97.9 & 91.1 & 88.4 & 75.7 & 77.1 & 94.3 & 83.8 \\
 & PGD & 4.8 & 10.9 & 5.2 & 4.8 & 13.8 & 9.2 & 4.1 \\
\midrule
\multirow{2}{*}{\rotatebox{0}{\textbf{ViT-L/14}}} & 
Clean & 99.4 & 94.2 & 90.9 & 79.0 & 80.1 & 95.9 & 88.7 \\
 & PGD & 6.8 & 15.7 & 12.8 & 7.7 & 14.5 & 21.2 & 8.0 \\

\bottomrule
\end{tabular}}}

\end{table}

In this section, we compare \emph{AdvPT} with linear prob CLIP, which also utilizes additional data, to investigate whether the robustness improvements of \emph{AdvPT} merely result from additional downstream data, as shown in \cref{tab:A1}.
By comparing with \cref{tab:1}, while it shows an increase in clean accuracy compared to vanilla CLIP, its robustness is reduced, when compared to \emph{AdvPT}.
This indicates that merely introducing additional downstream data does not directly contribute to enhanced robustness.
Furthermore, it also indicates that the enhancements in robustness are not entirely relevant to improvements in accuracy.

\subsection{Evaluation on Domain Shift}

\begin{figure}[t]
    \centering
    \begin{minipage}{0.48\textwidth}
        \centering
        \resizebox{1.0\linewidth}{!}{
        \setlength{\tabcolsep}{2.2mm}{
            \begin{tabular}{
              l
              l
              l
              S[table-format=2.1]
              S[table-format=2.1]
              S[table-format=2.1]
              S[table-format=2.1]
            }
            \toprule
            & & & {-V2} & {-A} & {-R} & {-Sketch} \\
            \midrule
            \multirow{4}{*}{\rotatebox{90}{ViT-B/16}} & \multirow{2}{*}{vanilla CLIP} & Clean & 60.8 & 47.7 & 80.5 & 46.9 \\
                                         & & Robust & 6.2 & 4.7 & 9.3 & 5.9 \\
            & \multirow{2}{*}{\emph{AdvPT}} & Clean & 62.6 & 46.3 & 83.6 & 45.6 \\
            &                              & Robust& 16.3 & 10.1 & 22.0 & 9.4 \\
            \midrule
            \multirow{4}{*}{\rotatebox{90}{ViT-L/14}} & \multirow{2}{*}{vanilla CLIP} & Clean & 67.9 & 68.7 & 91.8 & 57.2 \\
            &                             & Robust& 25.6 & 16.8 & 34.3 & 20.7 \\
            & \multirow{2}{*}{\emph{AdvPT}} & Clean & 71.1 & 69.0 & 92.1 & 58.5 \\
            &                            & Robust  & 38.5 & 20.2 & 43.5 & 25.8 \\
            \bottomrule
            \end{tabular}}}
        \caption{\emph{AdvPT} vs. vanilla CLIP on distribution shift.}
        \label{tab:3}
    \end{minipage}%
    \hspace{0.04\textwidth} 
    \begin{minipage}{0.4\textwidth}
        \centering
        \includegraphics[width=0.8\linewidth]{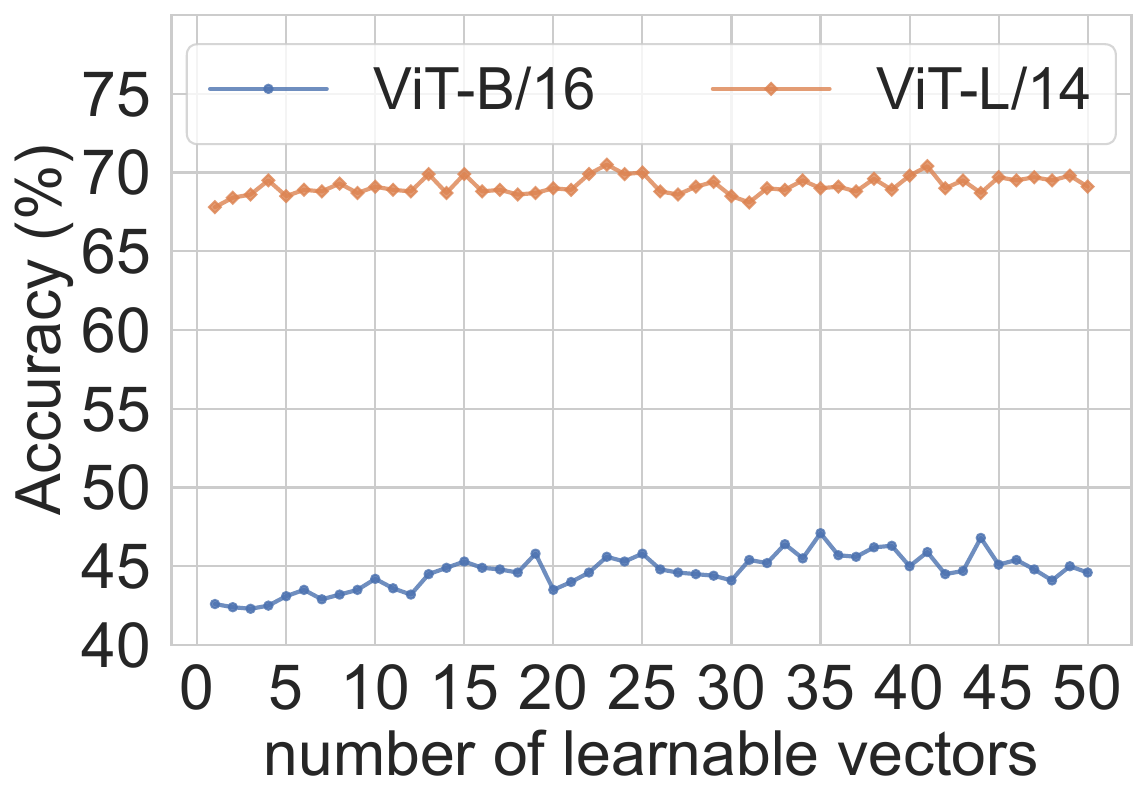} 
        \caption{Effect of number of learnable vector on Pets.} 
        \label{fig:4} 
    \end{minipage}
\end{figure}

A notable advantage of CLIP lies in its adaptability to domain shift. Thus, in this subsection, we evaluate the transferability of \emph{AdvPT} in comparison to the vanilla CLIP in domain shift scenarios. The source dataset utilized is ImageNet, while the target datasets include ImageNetV2, ImageNet-Sketch, ImageNet-A, and ImageNet-R. The results presented in \cref{tab:3} elucidate that the proposed \emph{AdvPT} outperforms the vanilla CLIP in terms of adversarial robustness, thereby validating its stability across varied domains.

\subsection{Further Analysis}


\subsubsection{Number of Learnable Vector}

In \cref{sec:5-4}, we observed similarities between adversarial prompt tuning and AT. 
It is widely acknowledged within the AT framework that a larger count of tunable parameters correlates with enhanced adversarial robustness. 
To discern whether this correlation persists within \emph{AdvPT}, we conducted an empirical evaluation of its efficacy under different numbers of learnable vector \( M \in [1, 50] \), using the Pets dataset as an example.
The empirical results, as illustrated in \cref{fig:4}, suggest that the volume of tunable parameters does not constitute a constraint in \emph{AdvPT}. 
Instead, unlocking its potential efficacy warrants further investigation.

\subsubsection{Interpreting the Learnable Vector}

In this subsection, we aim to decode what the learnable vectors have captured. 
However, direct mapping of these learnable vectors to words is infeasible due to the optimization occurring within a continuous space, while word space is discrete.
Therefore, we adopt a technique applied in the CoOp experiment, searching vocabulary for the nearest words to the learned vectors by Euclidean distance, as illustrated in \cref{tab:4}.
These words are not intuitively understandable, exactly aligning with the non-robust features in adversarial images.

\begin{table*}[t]
\centering
\caption{The nearest words for learnable vectors. N/A means non-Latin characters.}
\label{tab:4}
\resizebox{\textwidth}{!}{
\setlength{\tabcolsep}{0.2mm}{
\begin{tabular}{ccccccccc}
\toprule
     & \textbf{Flowers} & \textbf{Pets} & \textbf{Food101} & \textbf{SUN397} & \textbf{DTD} & \textbf{EuroSAT} & \textbf{UCF101} & \textbf{ImageNet} \\
    \hline
     & activated(0.6720) & stores(0.6300) & sii(1.6187) &  gaunt(1.4723) & 
     3(0.6263) & ust(0.8010) & laces(1.0643) & N/A(0.6407)\\

    & walked(0.7015) & sun(0.6388) & activation(1.6778) & maestro(1.5045) & 
     alization(0.6467) & trip(0.9385) & fa(1.1818) & le(0.6747)\\

     & pper(0.7994) & amore(0.6530) & thereal(1.6817) &  zoom(1.5162) & 
     cs(0.7361) & vu(1.0143) & deployed(1.2376 & telly(0.6995)\\

     & bao(0.8742) & favorites(0.6877) & cst(1.6910) &  nag(1.5209) & 
     prelude(0.7904) & salam(1.0190) & N/A(1.2625) & hooper(0.7082)\\
     
    & burden(0.8924) & ama(0.6957) & pancreatic(1.8803)  & cope(1.5922) & therapists(0.8336) & weymouth(1.1291) & cumbri(1.2966) & naq(0.7121)\\

\bottomrule
\end{tabular}}}
\end{table*}

\section{Limitations}

First, the paper's focus is restricted to image recognition tasks. Exploring the applicability of \emph{AdvPT} to a broader array of tasks, such as Visual Question Answering (VQA) in advanced models like GPT-4V~\cite{gpt}, is a worthwhile direction for future research.
Second, visual prompts~\cite{jia2022visual, khattak2023maple} emerge as a promising research avenue, given their extensive trainable parameters, which could enhance adversarial robustness. 
Yet, it introduces additional branches to the model, thus falling into the model robustification category.

\section{Conclusion and Discussion}

This study introduces Adversarial Prompt Tuning (\emph{AdvPT}), a novel technique enhancing the adversarial robustness of VLMs such as CLIP. Our approach, focusing on the alignment of learnable text prompts with adversarial image embeddings, represents a significant step forward in securing VLMs against adversarial attacks. Notably, \emph{AdvPT} achieves this heightened security without necessitating extensive model retraining or architectural modifications.

However, we acknowledge that this is an initial foray into a complex domain. 
Future research should explore the scalability of adversarial prompt tuning across various settings.
In conclusion, \emph{AdvPT} presents a promising direction for enhancing VLM's robustness, contributing to the broader endeavor of making AI systems more secure and reliable.

\section*{Acknowledgements}
This work was supported by National Key R\&D Program of China (Grant No. 2022ZD0160103, 2023YFC3310700), National Natural Science Foundation of China (Grant No. 62172094, 62276067) and Science and Technology Commission of Shanghai Municipality (Grant No. 22511106102).

\bibliographystyle{splncs04}
\bibliography{main}
\end{document}